%% file: 0_main.tex
\documentclass[sigconf]{acmart}
\AtBeginDocument{%
  }

\copyrightyear{2026}
\acmYear{2026}
\setcopyright{cc}
\setcctype{by-nc-nd}
\acmConference[WWW '26]{Proceedings of the ACM Web Conference 2026}{April 13--17, 2026}{Dubai, United Arab Emirates}
\acmBooktitle{Proceedings of the ACM Web Conference 2026 (WWW '26), April 13--17, 2026, Dubai, United Arab Emirates}
\acmPrice{}
\acmDOI{10.1145/3774904.3792139}
\acmISBN{979-8-4007-2307-0/2026/04}

\usepackage[utf8]{inputenc} 
\usepackage[T1]{fontenc}    
\usepackage{hyperref}       
 
\usepackage{url}            
\usepackage{booktabs}       
\usepackage{amsfonts}       
\usepackage{nicefrac}       
\usepackage{microtype}      
\usepackage[dvipsnames]{xcolor}         
\usepackage{enumitem}
\usepackage{amsthm}
\usepackage{amsmath}
\usepackage{bm}

\usepackage{amssymb}
\usepackage{mathtools}
\usepackage[linesnumbered,ruled,vlined]{algorithm2e}
\usepackage{makecell}
\usepackage{multirow}
\usepackage{subfigure} 
\usepackage{booktabs}
\usepackage{wrapfig}
\usepackage{extarrows}
\usepackage{colortbl}
\usepackage{tikz}
\usepackage{pifont}
\usepackage[most]{tcolorbox}
\usepackage{setspace}
\usepackage{adjustbox}

\newcommand{\modelname}{RAG-GFM}

\newcommand{\eg}{\textit{e.g.}}

\newcommand{\etc}{\textit{etc}}

\newcommand{\mycite}[1]{\hyperlink{cite.#1}{$\rhd$}}

\renewcommand\eqref[1]{\textup{(\ref{#1})}}

\theoremstyle{definition}

\newtheorem{prop}{Proposition}
\newtheorem{mydef}{Definition}

\newtcolorbox{mathbox}{
  colframe=black,
  colback=white,
  rounded corners,
  boxrule=0.5pt,
  breakable,
  left=1.5pt, right=1.5pt, top=1.5pt, bottom=1.5pt
}

\begin{document}

\title[RAG-GFM: Overcoming In-Memory Bottlenecks in Graph Foundation Models via Retrieval-Augmented Generation]{RAG-GFM: Overcoming In-Memory Bottlenecks in\\Graph Foundation Models via Retrieval-Augmented Generation}

\author{Haonan Yuan}
\affiliation{%
  \institution{SKLCCSE, School of Computer Science and Engineering \\ Beihang University}
  \city{Beijing}
  \country{China}
}
\email{yuanhn@buaa.edu.cn}

\author{Qingyun Sun}
\affiliation{%
  \institution{SKLCCSE, School of Computer Science and Engineering \\ Beihang University}
  \city{Beijing}
  \country{China}
}
\email{sunqy@buaa.edu.cn}

\author{Jiacheng Tao}
\affiliation{%
  \institution{SKLCCSE, School of Computer Science and Engineering \\ Beihang University}
  \city{Beijing}
  \country{China}
}
\email{jiachengtao@buaa.edu.cn}

\author{Xingcheng Fu}
\affiliation{%
  \institution{\makebox[0pt][c]{Key Lab of Education Blockchain and}\\ \makebox[0pt][c]{Intelligent Technology, Ministry of Education} \\ \makebox[0pt][c]{Guangxi Normal University}}
  \city{Guilin}
  \state{Guangxi}
  \country{China}
}
\email{fuxc@gxnu.edu.cn}

\author{Jianxin Li}
\authornote{Corresponding author.}
\affiliation{%
  \institution{SKLCCSE, School of Computer Science and Engineering \\ Beihang University}
  \city{Beijing}
  \country{China}
}
\email{lijx@buaa.edu.cn}


\begin{abstract}
  Graph Foundation Models (GFMs) have emerged as a frontier in graph learning, which are expected to deliver transferable representations across diverse tasks.
  However, GFMs remain constrained by \textit{\textbf{in-memory bottlenecks}}: they attempt to encode knowledge into model parameters, which limits semantic capacity, introduces heavy lossy compression with conflicts, and entangles graph representation with the knowledge in ways that hinder efficient adaptation, undermining scalability and interpretability.
  In this work, we propose \textbf{\modelname}, a \underline{\textbf{R}}etrieval-\underline{\textbf{A}}ugmented \underline{\textbf{G}}eneration aided \underline{\textbf{G}}raph \underline{\textbf{F}}oundation \underline{\textbf{M}}odel that offloads knowledge from parameters and complements parameterized learning.
  To externalize graph knowledge, we build a dual-modal unified retrieval module, where a semantic store from prefix-structured text and a structural store from centrality-based motif.
  To preserve heterogeneous information, we design a dual-view alignment objective that contrasts both modalities to capture both content and relational patterns.
  To enable efficient downstream adaptation, we perform in-context augmentation to enrich supporting instances with retrieved texts and motifs as contextual evidence.
  Extensive experiments on five benchmark graph datasets demonstrate that \modelname~consistently outperforms 13 state-of-the-art baselines in both cross-domain node and graph classification, achieving superior effectiveness and efficiency.
\end{abstract}





\maketitle

\input{1_intro}
\input{2_related_work}
\input{3_notation}
\input{4_method}
\input{5_experiment}
\input{6_conclusion}

\section*{Acknowledgments}
The corresponding author is Jianxin Li. Authors of this work are supported in part by NSFC under grants No.623B2010, No.62225202, and No.62302023, by the Fundamental Research Funds for the Central Universities, and by the Academic Excellence Foundation of BUAA for PhD Students. We extend our sincere thanks to all authors for their valuable contributions.

\newpage
\bibliographystyle{ACM-Reference-Format}
\bibliography{ref}

\newpage
\appendix
\counterwithin{table}{section}
\counterwithin{figure}{section}
\counterwithin{equation}{section}

\input{appendix/2_alg}

\input{appendix/3_proof}
\input{appendix/4_exp_settings}

\end{document}

%% file: 1_intro.tex
\section{Introduction}
\label{sec:intro}

Graphs are powerful structures for describing complex relationships among entities, and have been broadly adopted in domains such as modeling for the World Wide Web~\cite{ayers1995using, tadic2001dynamics}, social and citation networks~\cite{ fan2019graph, yuan2023environment}, retrieval and recommendation systems~\cite{wu2019session, wang2019knowledge}, knowledge graphs~\cite{wang2017knowledge, zhang2022knowledge}, biological analysis~\cite{gligorijevic2021structure, xiong2026multimodal}, \etc. Graph Neural Networks (GNNs)~\cite{hamilton2017inductive,kipf2022semi} have enabled effective representation learning on graphs, supporting a wide range of tasks, but are typically tailored to specific datasets and tasks, limiting cross-domain generalization. Motivated by large-scale pre-training in language and vision, Graph Foundation Models (GFMs) have recently emerged to learn universal graph representations through pre-training and downstream adaptation~\cite{mao2024position,shi2024graph,he2024unigraph,yuan2025graver,guo2025graphkeeper,shi2026sa,luo2026towards}, aiming to support diverse applications with minimal supervision.

Despite these advances, existing GFMs face fundamental limitations. Current methods follow the ``pretrain-then-finetune’’ paradigm overwhelmingly, where knowledge from source domains is either fully compressed into a single GFM’s model parameters~\cite{zhao2024all, yu2025samgpt}, or at best expanded through lightweight mixtures-of-experts (MoE) that remain largely conceptual and offer little practical relief~\cite{guo2024graphmore, yuan2025much}. While parameter counts may increase marginally, they fall far short of matching the vast, orders-of-magnitude greater knowledge volume inherent in pre-training domains. Graph knowledge is inherently dual-modal, combining node-level semantic texts and higher-order structural patterns, which leads to inevitable \textit{\textbf{in-memory bottlenecks}} that hinder scalability, robustness, and interpretability.


\begin{figure}[!t]
    \centering
    \includegraphics[width=1\linewidth]{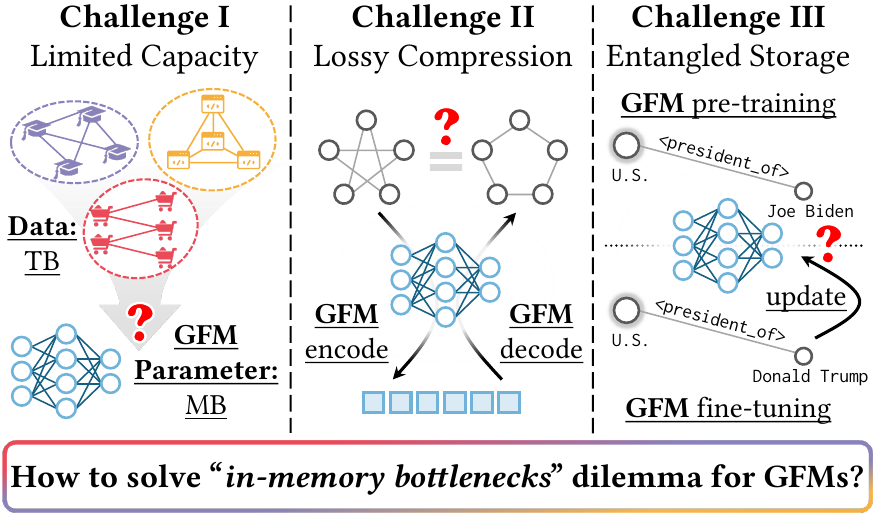}
    \vspace{-0.5cm}
    \caption{Challenges of the ``\textit{in-memory bottlenecks}''.}
    \label{fig:intro_compare}
    \vspace{-0.3cm}
\end{figure}

\textbf{Challenge I: Limited capacity within parameters.}  
Graph knowledge spans rich information whose scale far exceeds what model parameters can store. In graph models, increasing parameters or depth often causes over-smoothing rather than higher capacity~\cite{chen2020measuring,keriven2022not}. Consequently, GFMs trained on a single domain quickly exhaust their parameter budget when transferred to others with distinct semantics and motifs, leading to forgetting, poor transfer, and limited scalability~\cite{ramasesh2021effect,zhao2024continual}. This exposes the fundamental limitation of parameter-centric storage for graphs.

\textbf{Challenge II: Lossy and conflicting compression.}  
Forcing heterogeneous graph knowledge into parameters inevitably causes lossy and conflicting compression. Identical structural patterns may carry opposite semantics across domains, and collapsing them into shared embeddings distorts meaning. Moreover, such compression is irreversible: once absorbed into weights, knowledge cannot be retrieved, verified, or updated without retraining, undermining transparency and grounded reasoning.

\textbf{Challenge III: Entangled representation and storage.}  
The parameter-based storage tightly entangles knowledge with representations, hindering efficient adaptation. Fine-tuning simultaneously adjusts task-specific features and updates memorized knowledge, making learning inefficient and data-intensive. This entanglement also obscures interpretability, as predictions cannot be traced to explicit evidence, reducing reliability in high-stakes applications.


\textbf{Our key insight} is to move beyond parameter-centric storage by externalizing graph knowledge, inspired by Retrieval-Augmented Generation (RAG)~\cite{lewis2020retrieval}. Unlike text, graph knowledge is fragmented across attributes and structures, making retrieval more challenging. Existing GFMs compress such evidence into parameters, losing explicit access and updatability. We argue that treating graph knowledge as first-class retrievable objects enables external storage, aligned pre-training, and scalable, interpretable adaptation.


In this work, we propose \textbf{\modelname}, a \underline{\textbf{R}}etrieval-\underline{\textbf{A}}ugmented \underline{\textbf{G}}eneration aided \underline{\textbf{G}}raph \underline{\textbf{F}}oundation \underline{\textbf{M}}odel.  
\modelname~incorporates three key components.  
First, we construct a dual-store unified retrieval database, consisting of a semantic store from prefix-structured text embeddings and a structural store from centrality-based motif encodings, enabling GFMs to query external knowledge on demand.  
Second, we design a dual-view alignment objective that contrasts semantic representation with structural subgraphs during pre-training, ensuring complementary representation learning across modalities.  
Third, we introduce in-context sample augmentation, where retrieved texts and motifs are appended as contextual evidence for few-shot adaptation, enriching support instances with explicit external knowledge.  
\textbf{Our contributions are:}
\vspace{0.15cm}
\begin{itemize}[leftmargin=1em]
    \item We propose \modelname, the first retrieval-augmented graph foundation model that explicitly addresses in-memory bottlenecks.  
    \item We design a dual-store retrieval module, a dual-view alignment objective, and an in-context sample augmentation mechanism, providing a unified pipeline for knowledge externalization, robust pre-training, and efficient adaptation.  
    \item Extensive experiments on six benchmark graph datasets demonstrate that \modelname~consistently outperforms 13 state-of-the-art GFM baselines in both cross-domain node and graph classification, achieving superior effectiveness and efficiency.  
\end{itemize}

%% file: 2_related_work.tex
\section{Related Work}
\label{sec:related_work}

\textbf{Graph Foundation Models (GFMs).}
GFMs extend large-scale pre-training to graphs via self-supervised learning~\cite{sun2022gppt,yang2022self,chen2022pre,cao2023pre,yin2023train,luo2026privacy}. Most assume distributional similarity between pre-training and downstream tasks~\cite{huang2022few,zhang2021motif}, limiting robustness under domain shift. Recent work explores cross- or multi-domain learning, LLM alignment, domain tokens, and structural guarantees~\cite{zhao2024all,liu2024one,yi2023contrastive,tang2024cross,he2024unigraph,liu2024one,tang2024higpt,yu2024text,wang2024can,tan2024walklm,yuan2025much,wang2025multi}, yet GFMs remain parameter-centric and struggle with structural and semantic consistency.

\textbf{Retrieval-Augmented Generation (RAG).}
RAG enhances the LLMs by retrieving external knowledge to mitigate context limits and hallucinations~\cite{fan2024survey}. Using lexical or semantic retrieval with query optimization and re-ranking~\cite{ram2023context,ma2023query,li2023structure}, RAG achieves strong performance in QA and reasoning~\cite{tang2024multihop,jiang2023active}. Extensions to graph data motivate GraphRAG~\cite{xia2021graph,ma2021deep,han2024retrieval,peng2024graph,tian2024graph,kim2023factkg,sun2025dyg}. While GFM-RAG~\cite{luo2025gfm} uses GFMs to improve RAG, we instead leverage RAG to fundamentally enhance GFMs.

%% file: 3_notation.tex
\section{Notations and Preliminaries}
\label{sec:notation}

\textbf{Notations.}
We represent a graph as $G = (\mathcal{V}, \mathcal{E})$, where $\mathcal{V}$ denotes the set of nodes and $\mathcal{E}$ the set of edges. For a graph $G_i$ sampled from any of the source domains, let $\mathbf{A}\in\{0,1\}^{N_i \times N_i}$ be the adjacency matrix and $\mathbf{X}\in\mathbb{R}^{N_i \times d_i}$ be the node feature matrix. Here, $N_i = |\mathcal{V}_i|$ denotes the number of nodes, and $d_i$ denotes the original input feature dimension. $\mathbf{Z}$, $\mathbf{W}$, $\mathbf{H}$ are the hidden representations.

\textbf{Multi-domain Pre-training.}
Let $\mathcal{G}^\mathcal{S}=\{G_1^\mathcal{S},\cdots,G_n^\mathcal{S}\}$ denote a collection of source graphs from domains $\mathcal{D}^\mathcal{S}$, each associated with labels $\mathcal{Y}^\mathcal{S}$. 
We cast pre-training as a self-supervised link prediction problem with universal templates~\cite{liu2023graphprompt}, ensuring task consistency with downstream settings. 
The learner is defined as $h=g(f_{\boldsymbol{\Theta}}(\cdot))$, where the encoder $f\!:\mathbb{R}^{d_i}\mapsto\mathbb{R}^d$ produces node embeddings and the discriminator $g\!:\mathbb{R}^d\times\mathbb{R}^d\mapsto\mathbb{R}^2$ predicts link existence. 
Once the pre-training converges, parameter $\boldsymbol{\Theta}^\star$ is frozen as the backbone.

\textbf{Few-shot Fine-tuning.}
We consider graphs $\mathcal{G}^\mathcal{T}$ from target domains $\mathcal{D}^\mathcal{T}$ (seen or unseen). 
Under $m$-shot setting ($m \ll \sum_{i=1}^n N_i$), only $m$ labeled samples $\mathcal{Y}^\mathcal{T}$ are available. 
Fine-tuning applies the pre-trained $h=g(f_{\boldsymbol{\Theta}}^\star(\cdot))$ augmented with learnable prompts $\boldsymbol{\mathcal{P}}_{\boldsymbol{\Omega}}$, where $\boldsymbol{\Omega}$ denotes tunable parameters. 
Both node and graph classification (via node-centered ego-graphs) are reformulated as link prediction, maintaining homogeneity with pre-training.

%% file: 4_method.tex
\begin{figure*}[!t]
    \centering
        \begin{minipage}{\textwidth}
            \hspace{-0.1cm}
            \begin{adjustbox}{width=1.01\linewidth}
            \input{fig/framework_marked}
            \end{adjustbox}
        \end{minipage}
    \vspace{-0.3cm}
    \caption{The framework of \modelname. The framework includes three stages: (1) Unified Semantic-Structural Bi-Modal Retrieval Database for externalizing graph knowledge, (2) Cross-View Knowledge Alignment for pre-training transferable priors, and (3) In-Context Retrieval Augmentation for few-shot adaptation via domain-gated lightweight graph prompts.}
\label{fig:framework}
\vspace{-0.5em}
\end{figure*}
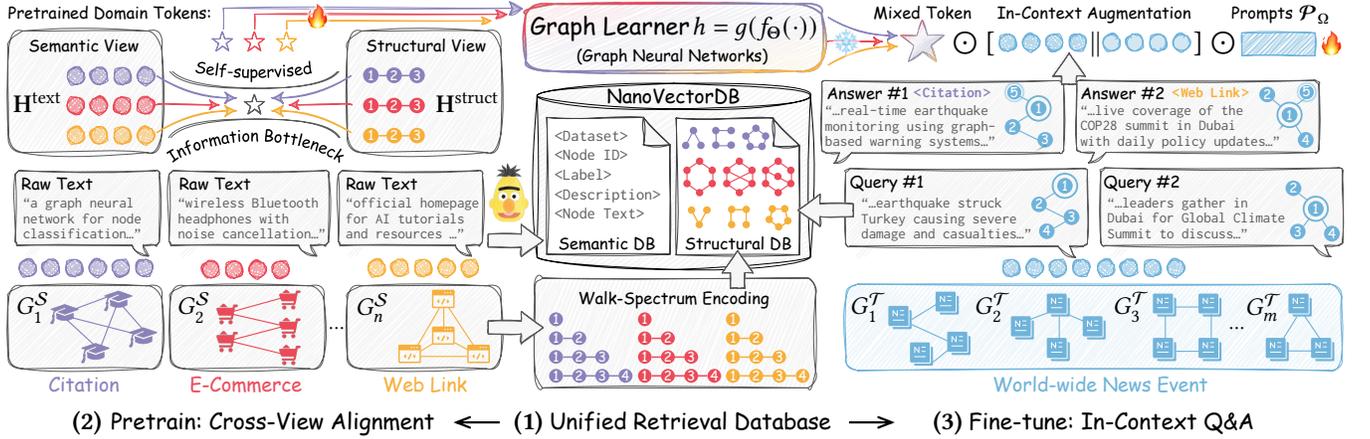

\section{Method}
\label{sec:method}
We illustrate the framework of \modelname~in Figure~\ref{fig:framework}.

\subsection{Unified Semantic-Structural Bi-Modal Retrieval Database}
\label{sec:db}
At the core of our framework lies a unified retrieval database that externalizes knowledge into two complementary modalities: a \textit{\textbf{semantic store}} that organizes enriched node texts as retrievable documents, and a \textit{\textbf{structural store}} that captures motif-level patterns.
Formally, we denote the bi-modal database\footnote{Implemented in \texttt{NanoVectorDB}~\cite{ye2024nano}, a lightweight but efficient database that provides scalable storage and top-$k$ retrieval over dense queries.} as $\mathcal{D}\!=\!\{\mathcal{D}_\text{text}, \mathcal{D}_\text{struct}\}$.
Given query $\mathbf{q}$ and scoring function $s(\cdot,\cdot)$, the retrieval operator is:
\begin{equation}
    \mathrm{Retrieve}(\mathcal{D}, \mathbf{q}, k) = \arg \text{Top-}k_{(u,\mathbf{z}_u) \in \mathcal{D}}\left[s(\mathbf{q}, \mathbf{z}_u)\right],
\end{equation}
with each entry $(u,\mathbf{z}_u)$ denoting a database record identified by $u$ and described by $\mathbf{z}_u$. $\mathcal{D}$ is queried in both pre-training and fine-tuning, which allows \modelname~to ground predictions in explicit semantic and structural evidence rather than obscure parameters.

\subsubsection{\textbf{Semantic Store.}}
\label{sec:initialization}
Unlike raw node features reduced to numerical vectors, most of the graphs are text-attributed, with nodes from sources such as abstracts, product descriptions, or profiles. We recover each node’s raw text ${\mathbf{t}_v}$ by tracing it back to its original corpus (\eg, metadata in citations)~\cite{liu2023one, li2024zerog}, and treat it as a first-class signal.
The semantic pipeline branches into two tracks:

On the \textit{\textbf{representation track}}, we address the dimension-wise mismatch of raw features across domains using PCA~\cite{pearson1901liii}.
For each graph $G_i$ with feature matrix $\mathbf{X}_i \in \mathbb{R}^{N_i \times d_i}$, we apply:
\begin{equation}
    \widetilde{\mathbf{X}}_i^\mathcal S = \mathrm{PCA}_{d_0}(\mathbf{X}_i^\mathcal S) \in \mathbb{R}^{N_i \times {d_0}}.
\label{eq:align}
\end{equation}
In parallel, the raw text $t_v$ is encoded by a BERT into a semantic vector $\mathbf{b}_v \in \mathbb{R}^{d_0}$.
The updated node feature is $\widehat{\mathbf{x}}_v^\mathcal S = \left[\widetilde{\mathbf{x}}_v^\mathcal S ~\|~ \mathbf{b}_v \right] \in \mathbb{R}^{2d_0}$, combining dimension-aligned attributes with enriched semantics.
Thus, we learn the graph embeddings with the text-wise encoder:
\begin{equation}
    \mathbf{Z}_i^\mathcal S = f_{\boldsymbol{\Theta}_t}\big(\widehat{\mathbf{X}}_i^\mathcal S, \mathbf{A}_i^\mathcal S\big) \in \mathbb{R}^{N_i \times d}.
\label{eq:semantic_view}
\end{equation}

On the \textit{\textbf{retrieval track}}, we build $\mathcal{D}_{\text{text}}$ as a textual vector database. For each node and its corresponding raw text, to standardize heterogeneous sources and make retrieval controllable and interpretable, we augment each document with a structured prefix:

\begin{tcolorbox}[
    enhanced,
    colback=white,
    colframe=black!80,
    boxrule=0.7pt,
    rounded corners,
    fonttitle=\bfseries,
    title=Prefix Schema \qquad~\qquad~\qquad~\qquad~\qquad~\qquad~\qquad~\qquad~\qquad~\qquad~\qquad~\qquad~\qquad~\qquad~\qquad~\qquad~\qquad~\qquad Example,
    sidebyside,
    sidebyside align=top seam,
    lefthand width=3.75cm,
    boxsep=2pt,
    left=2pt,
    right=2pt,
    top=3pt,
    bottom=-7.2pt,
    sidebyside gap=10pt
]

\begin{spacing}{0.88}
{\small 
\texttt{Dataset: <dataset\_name>}\\
\texttt{Node ID: <node\_id>}\\
\texttt{Label: <node\_label>}\\
\texttt{Description: <description>}\\
\texttt{Node Text: <node\_text>}
}
\end{spacing}

\tcblower

\begin{spacing}{0.88}
{\small
\textit{Cora}\\
\#123\\
\textit{Neural Networks}\\
\textit{Papers about neural networks.}\\
\textit{This paper introduces the LSTM, a Long Short-Term Memory model.}
}
\end{spacing}
\end{tcolorbox}

\noindent The prefixed document $\widetilde{\mathbf{t}}_v$ is segmented not by naive length rules but into graph-aware chunks $\{\mathbf{c}_{v_1},\cdots,\mathbf{c}_{v_k}\}$ aligned with descriptive fields and class-level information for fine-grained retrieval. In this way, it yields coherent chunks that remain intrinsically aligned with the structure rather than arbitrary spans. Each chunk is embedded with BERT into $\widetilde{\mathbf{z}}^\mathcal{S}_{v_j}\in\mathbb{R}^{768}$. We insert $(v,\widetilde{\mathbf{z}}^\mathcal{S}_{v_j},\text{meta}_{v_j})$ into the $\mathcal{D}_{\text{text}}$, where $\text{meta}_{v_j}$ carries the structured fields from the prefix. Formally,
\begin{equation}
    \mathcal{D}_\text{text}=\big\{\big(v,\widetilde{\mathbf{z}}^\mathcal{S}_{v_j},\text{meta}_{v_j}\big)\mid v\in\{\mathcal{V}_i\}_{i=1}^n, j\in[1,k]\big\}.
\label{eq:retrieval_track}
\end{equation}
The prefix serves as a ``retrieval hook'' for the metadata filtering and cross-domain alignment, while the 768-dimensional embeddings preserve semantic capacity for in-context augmentation.

\subsubsection{\textbf{Structural Store.}}
Enumerating motifs is computationally intractable (NP-hard), and storing arbitrary subgraphs introduces noise.
Inspired by~\cite{estrada2005subgraph, southern2025balancing}, we propose the \underline{\textbf{W}}alk-\underline{\textbf{S}}pectrum \underline{\textbf{E}}ncoding (\textbf{WSE}), which ranks nodes by a walk-based importance and encodes their local neighborhoods with a multi-order walk signature.
\begin{mathbox}
    \begin{mydef}[\textbf{Walk-Spectrum Encoding}]
        For a node $v \in \mathcal{V}$, the Walk-Spectrum Encoding (WSE) of order $K$ is defined as:
        \begin{equation}
            \mathbf{C}^{\text{WSE}}_\alpha(v) = \big[ \alpha\mathbf{A}_{vv},\ \alpha^2\mathbf{A}^2_{vv},\ \alpha^3\mathbf{A}^3_{vv}, \cdots, \alpha^K\mathbf{A}^K_{vv} \big],
        \end{equation}
        where $\alpha\!\in\!(0,1)$ is a damped variant, and $\mathbf{A}^k_{vv}$ counts the number of closed walks of length $k$ starting and ending at node $v$.
    \end{mydef}
\end{mathbox}
\noindent WSE summarizes a node’s participation in closed walks of varying lengths, thereby encoding structural patterns beyond any fixed-radius neighbors. This motivates the following result on its ability to separate graphs that local methods~\cite{southern2025balancing} cannot distinguish:

\begin{mathbox}
    \begin{prop}[\textbf{Structural Separability of WSE}]
    \label{prop:wse}
        There exist pairs of non-isomorphic graphs $G_1$, $G_2$ and nodes $v\in G_1$, $u\in G_2$ such that for any fixed radius $r$, the $r$-hop neighbors $\mathcal{N}_r(v)$ and $\mathcal{N}_r(u)$ are isomorphic, yet the Walk-Spectrum Encodings satisfy:
        \begin{equation}
            \mathbf{C}^{\text{WSE}}_\alpha(v) \ne \mathbf{C}^{\text{WSE}}_\alpha(u).
        \end{equation}
    \end{prop}
\end{mathbox}
\noindent Proofs in Appendix~\ref{app:proof1}.
While WSE provides rich structural signatures, computing and storing subgraphs for all nodes is infeasible at scale. To address this, we derive an anchor scoring function:
\begin{equation}
    r_v(\alpha,K)=\sum\nolimits_{k=1}^{K} \alpha^k \mathbf{A}^k_{vv},\quad\text{for each}~v\in\{\mathcal{V}_i\}_{i=1}^n
\end{equation}
Intuitively, $r_v$ highlights nodes most recurrently involved in structural motifs. Ranking nodes by $r_v$ allows us to select a compact yet informative pool of anchors for motif storage and retrieval.

We then select the top-$M$ nodes in each graph into $\mathcal{V}^\mathcal{S}_\text{anchor}$, and extract its $h$-hop ego-subgraph $G_{v(h)}^\mathcal{S}$ for each node $v$. To reduce redundancy, overlapping ego-subgraphs are pruned via node set equivalence. The resulting structural store is defined as:
\begin{equation}
    \mathcal{D}_\text{struct}=\big\{\big(v,G_{v(h)}^\mathcal{S},\mathbf{C}^{\text{WSE}}_\alpha(v),\text{meta}_{v}\big)\mid v\in\mathcal{V}^\mathcal{S}_\text{anchor}\big\},
\label{eq:structual_store}
\end{equation}
where $\text{meta}_{v}$ includes metadata like hop radius, anchor score, \etc.
At this point, we have established the unified semantic-structural bi-modal retrieval database $\mathcal{D}$, which will serve as the foundation for subsequent pre-training and fine-tuning.

\subsection{Cross-View Knowledge Alignment for Multi-domain Pre-training}
\label{sec:align}
With the unified database $\mathcal{D}=\{\mathcal{D}_{\text{text}},\mathcal{D}_{\text{struct}}\}$ in place, the next step is to pre-train the encoder $f_{\boldsymbol{\Theta}}(\cdot)$ that couples semantic and structural information in a principled way. The goal is not to collapse them into a single representation but to ensure that both carry complementary and transferable signals across domains.

\subsubsection{\textbf{Node Views.}}
For each $G_i^{\mathcal S}$, we build two node-level views. The semantic view is given by the enriched node embeddings $\mathbf{Z}_i^{\mathcal S}$ in Eq.~\eqref{eq:semantic_view}, which combine raw attributes and text-derived features. The structural view is constructed from the walk-spectrum encoding:
\begin{equation}
    \mathbf{W}_i^{\mathcal S}=\big[\mathbf{C}^{\text{WSE}}_\alpha(v)\big]_{v \in \mathcal{V}_i^{\mathcal S}},
\label{eq:strcutural_view}
\end{equation}
where each item records closed-walk signatures up to order $K$, capturing recurring motif patterns and higher-order relational signals.

\textbf{Domain Tokens.}
To incorporate domain-level priors, we introduce a learnable token $\boldsymbol{\tau}_{D_i}\!\in\!\mathbb{R}^{d_{\boldsymbol{\tau}}}$ for each source domain $D_i^{\mathcal S}$, which is concatenated to every node representation before encoding:
\begin{equation}
    \overline{\mathbf{Z}}~\!_i^{\mathcal S} = \big[\mathbf{Z}_i^{\mathcal S}\ \big\|\ \mathbf{1}\cdot\boldsymbol{\tau}_{D_i}^\top\big], \qquad \overline{\mathbf{W}}~\!_i^{\mathcal S} = \big[\mathbf{W}_i^{\mathcal S}\ \big\|\ \mathbf{1}\cdot\boldsymbol{\tau}_{D_i}^\top\big],
\end{equation}
where $\mathbf{1}$ denotes a broadcast vector ensuring nodes within a domain share this token. During optimization, $\boldsymbol{\tau}_{D_i}$ accumulates domain priors that are not captured by individual nodes, such as global semantics in citation graphs or biochemical motifs in protein-protein networks. Tokens initialize lightweight graph prompts at fine-tuning, enabling adaptation without revisiting the full pre-training corpus.

\subsubsection{\textbf{Cross-View Information Bottleneck}}
Our pre-training is entirely self-supervised: the key idea is to align semantic and structural views of the same node without relying on labels, while simultaneously preventing collapse by encouraging each view to preserve modality-specific information.
We apply two encoders over the same topology but different features:
\begin{equation}
    \mathbf{H}_i^{\text{text}}=f_{\boldsymbol{\Theta}_t}\big(\overline{\mathbf{Z}}~\!_i^{\mathcal S},\mathbf{A}_i^{\mathcal S}\big), \qquad \mathbf{H}_i^{\text{struct}}=f_{\boldsymbol{\Theta}_s}\big(\overline{\mathbf{W}}~\!_i^{\mathcal S}, \mathbf{A}_i^{\mathcal S}\big),
\label{eq:pretrain_encode}
\end{equation}
which yields semantic embeddings $\mathbf{h}^{\text{text}}_{i,v}$ and structural embeddings $\mathbf{h}^{\text{struct}}_{i,v}$ for each node $v$.
Concretely, we introduce the self-supervised information bottleneck~\cite{tishby2015deep} by maximizing the mutual information between semantic and structural embeddings, and applying compression regularizers to discard redundant signals:
\begin{equation}
    \mathcal{L}_{\text{align}}^{(i,v)} = \underbrace{- I\Big(\mathbf{h}^{\text{text}}_{i,v};\mathbf{h}^{\text{struct}}_{i,v}\Big)}_{\text{\normalsize relevance}}+\beta\underbrace{\Big(I\big(\mathbf{h}^{\text{text}}_{i,v};\overline{\mathbf{z}}~\!_{i,v}^{\mathcal S}\big) + I\big(\mathbf{h}^{\text{struct}}_{i,v};\overline{\mathbf{w}}~\!_{i,v}^{\mathcal S}\big)\Big)}_{\text{\normalsize compression}},
\label{eq:intractbale}
\end{equation}
where $I(\cdot;\cdot)$ denotes the mutual information, which is intractable over unknown latent distributions of variables, and $\beta$ is the trade-off hyper-parameter. We adopt a contrastive approximation that yields a variational bound for tractable computation~\cite{tishby2000information, kraskov2004estimating, tishby2015deep, sun2026information}:
\begin{mathbox}
    \begin{prop}[\textbf{Cross-View Mutual Information Bounds}]
    \label{prop:ib}
    The relevance term admits the InfoNCE lower-bound estimator:
        \begin{equation}
            \!\!\!\!\!I\big(\mathbf{h}^{\text{text}}_{i,v};\mathbf{h}^{\text{struct}}_{i,v}\big) \!\!\geqslant\!\! \frac{1}{|\mathcal{B}|}\!\sum_{v\in\mathcal{B}}\!\log\!\frac{\exp\!\big(\sigma\big(g_t(\mathbf{h}^{\text{text}}_{i,v}),g_s(\mathbf{h}^{\text{struct}}_{i,v})\big)\!/\!\tau\big)}{\sum\limits_{u\in\mathcal{B}}\!\!\exp\!\big(\sigma\big(g_t(\mathbf{h}^{\text{text}}_{i,v}),g_s(\mathbf{h}^{\text{struct}}_{i,u})\big)\!/\!\tau\big)},\!\!\!
        \label{eq:lb}
        \end{equation}
        where $g_t$, $ g_s$ are projections, $\sigma(\cdot)$ is similarity, $\tau$ is a temperature, positives are formed by the same node across the views $(v,v)$ in a batch $\mathcal{B}$, and negatives by mismatched nodes $(v,u), u \!\neq\! v$.
        \vspace{0.1cm}
        \noindent\makebox[\linewidth]{\dotfill}
        The compression term can be upper-bounded via KL-divergence:
        \begin{equation}
            \!\!\!\!I\big(\mathbf{h}_{i,v}^\cdot;\overline{\mathbf{x}}~\!_{i,v}^{\mathcal S}\big) \!\!\leqslant\! \!\mathbb{E}_{p(\mathbf{h}_{i,v}^\cdot,\overline{\mathbf{x}}~\!_{i,v}^{\mathcal S})} \!\big[\!\log q_\phi\big(\mathbf{h}_{i,v}^\cdot|\overline{\mathbf{x}}~\!_{i,v}^{\mathcal S}\big)\!\big] \!-\! \mathbb{E}_{p(\mathbf{h}_{i,v}^\cdot)}\! \big[\!\log p\big(\mathbf{h}_{i,v}^\cdot\big)\!\big],\!\!
        \label{eq:ub}
        \end{equation}
        where $v$ is sampled from $\mathcal{B}$, $\mathbf{x}$ denotes $\mathbf{z}$ or $\mathbf{w}$, and $q_\phi(\cdot|\cdot)$ is a variational approximation of the true conditional distribution.
    \end{prop}
\end{mathbox}
\noindent Proposition~\ref{prop:ib} provides tractable self-supervised estimators for the otherwise intractable mutual information terms, with lower bounds applied to cross-view alignment and upper bounds applied to view-specific compression. We provide sketch proofs in Appendix~\ref{app:proof2}.

\subsubsection{\textbf{Pre-training Objective}}
Bringing the above components together, the overall pre-training objective is defined as:
\begin{equation}
    \mathcal{L}_{\text{pretrain}}(\boldsymbol{\Theta}_t,\boldsymbol{\Theta}_s) =\sum_{D_i^{\mathcal S}} \frac{1}{\left|\mathcal{V}_i^{\mathcal S}\right|}\sum_{v\in \mathcal{V}_i^{\mathcal S}}\mathcal{L}_{\text{align}}^{(i,v)}\cdot \gamma \sum_{D_i^{\mathcal S}}\left\|\boldsymbol{\tau}_{D_i}\right\|_2^2,
\label{eq:pretrain_loss}
\end{equation}
where the first term aggregates the cross-view alignment loss across source domains, and the second term regularizes domain tokens to prevent overfitting. $\gamma$ acts as their trade-off hyper-parameter.

In practice, mini-batches are constructed by mixing nodes from different domains, and the corresponding domain tokens are updated jointly with semantic and structural encoders. This setup enforces cross-domain consistency during pre-training while preserving domain-specific priors for downstream adaptation. $\boldsymbol{\Theta}_t$ and $\boldsymbol{\Theta}_s$ are frozon once pre-training converges.
We illustrate the pre-training pipeline in Algorithm~\ref{alg:pretrain} with its complexity analysis.

\subsection{In-Context Retrieval Augmentation for Few-Shot Fine-Tuning}
\label{sec:qa}
We proceed to fine-tune pre-trained model under \textit{\textbf{meta-learning}} settings ($m$-shot), which is more challenging in real-world scenarios.

\subsubsection{\textbf{Domain-Gated Fusion.}}
To ensure dimension-consistency with pre-training, each support sample is processed by the same representation track (Section~\ref{sec:initialization}, Eq.~\eqref{eq:align} and Eq.~\eqref{eq:semantic_view}) into $\widehat{\mathbf{X}}^{\mathcal T}$.

Before retrieval, we estimate domain affinities that will route external evidence. For each target node $v_i$ or graph $G_i^\mathcal{T}$, we compute soft gating weights over source domains via domain tokens:
\begin{equation}
    \pi_{i,k} = \frac{\exp\big(\sigma(\mathbf{Z}_{i}^\mathcal{T},\boldsymbol{\tau}_{D_k})\big)}{\sum_j \exp\big(\sigma(\mathbf{Z}_{i}^\mathcal{T},\boldsymbol{\tau}_{D_j})\big)},\quad \mathbf{Z}_i^\mathcal T = f_{\boldsymbol{\Theta}_t}^\star\big(\widehat{\mathbf{X}}_i^\mathcal T, \mathbf{A}_i^\mathcal T\big),  
\label{eq:gate}
\end{equation}
where $\mathbf{Z}_{i}^\mathcal{T}$ is the pre-trained encoder output on $\widehat{\mathbf{X}}^{\mathcal T}$. These $\{\pi_{i,k}\}_{k=1}^n$ act as domain-aware gates reused by the later augmentations.

\subsubsection{\textbf{Query and Retrieval.}}
\label{sec:qanda}
For clarity, we present query and retrieval in node-level settings. The extension to graph-level tasks follows directly by treating each graph as a single instance.

\textbf{Semantic Retrieval.}
For each few-shot target node $v$ and its one-hop neighbors, we form a \textit{\textbf{textual query}} $\mathbf{q}_v^\text{text}$ from the raw text and submit it to $\mathcal{D}_{\text{text}}$, restricting search to pre-training domains to avoid leakage. The database returns top-$k$ \textit{\textbf{textual answers}} $\{\widetilde{\mathbf{z}}^\mathcal{S}_{u}\}$, which are aggregated through softmax-weighted fusion:
\begin{equation}
    \big(\Delta\mathbf{z}^{\mathcal{T}}_v\big)^\text{text} = \!\!\!\sum_{u \in \text{Top-}k(v)}\!\!\! w_{vu}\cdot \widetilde{\mathbf{z}}^\mathcal{S}_{u},\quad w_{vu} = \frac{\exp\big(\sigma(\mathbf{q}_v^\text{text},\widetilde{\mathbf{z}}^\mathcal{S}_{u})\big)}{\sum_{u^\prime} \exp\big(\sigma(\mathbf{q}_v^\text{text},\widetilde{\mathbf{z}}^\mathcal{S}_{u^\prime})\big)},
\label{eq:text_query}
\end{equation}
where $\mathbf{z}^{\mathcal T}_v$ is in-context augmented with hyper-parameter $\lambda_\text{text}$:
\begin{equation}
    \mathbf{z}^{\mathcal{T}\prime}_v = \mathbf{z}^{\mathcal T}_v + \lambda_\text{text}\cdot \big(\Delta\mathbf{z}^{\mathcal{T}}_v\big)^\text{text}.
\label{eq:augment_1}
\end{equation}

\textbf{Structural Retrieval.}
For the same node $v$ and its neighbors, we extract the $h$-hop subgraph, encode it with WSE as the \textit{\textbf{structural query}}, and submit $\mathbf{q}_v^\text{struct}$ to $\mathcal{D}_{\text{struct}}$. From each source domain $D_i^{\mathcal S}$, we retrieve the most structurally similar motif $G_{v(h),i}^\mathcal{S}=\big(\widehat{\mathbf{X}}_{v(h),i}^\mathcal{T},\mathbf{A}_{v(h),i}^\mathcal{T}\big)$ as \textit{\textbf{strcutural answer}}. We then fuse cross-domain answers using domain gates $\{\pi_{v,k}\}$ with hyper-parameter $\lambda_\text{struct}$:
\begin{align}
    &\mathbf{z}^{\mathcal{T}\prime\prime}_v = \mathbf{z}^{\mathcal{T}\prime}_v + \lambda_\text{struct}\cdot \big(\Delta\mathbf{z}^{\mathcal{T}}_v\big)^\text{struct},
    \label{eq:augment_2}
    \\
    &\big(\Delta\mathbf{z}^{\mathcal{T}}_v\big)^\text{struct}=\sum\nolimits_{D_i^\mathcal{S}}\pi_{v,k} \big(f_{\boldsymbol{\Theta}_s}^\star\big(\widehat{\mathbf{X}}_{v(h),i}^\mathcal{T},\mathbf{A}_{v(h),i}^\mathcal{T}\big)\big).
    \label{eq:struct_query}
\end{align}

\subsubsection{\textbf{Prompted Few-shot Adaptation.}}
Given $m$ retrieved and augmented support samples $\{(\mathbf{h}_i^\mathcal{T},\mathbf{y}_i)\}$, to enable efficient adaptation without updating the frozen $\boldsymbol{\Theta}_t$ and $\boldsymbol{\Theta}_s$, we initialize learnable graph prompts $\boldsymbol{\mathcal{P}_\Omega}$ by the routed domain priors:
\begin{equation}
    \mathbf{h}_i^\mathcal{T} = \big[\mathbf{z}^{\mathcal{T}\prime\prime}_i \| \boldsymbol{\mathcal{P}_\Omega}\big],\quad \boldsymbol{\mathcal{P}_\Omega} \gets \sum\nolimits_{k=1}^n \pi_{i,k} \boldsymbol{\tau}_{D_k},
\label{eq:prompt}
\end{equation}
where $\mathbf{h}_i^\mathcal{T}$ denotes the $i$-th target node or graph embedding.
The fine-tuning objective is transformed into determining the similarity between the query sample and the class prototype embedding:
\begin{equation}
    \!\!\!\mathcal{L}_\text{fine-tune}(\boldsymbol{\mathcal{P}_\Omega})=-\!\!\!\!\!\sum_{\{(\mathbf{h}_i^\mathcal{T},\mathbf{y}_i)\}}\!\left[\log\frac{\exp\big(g(\mathbf{h}_i^\mathcal{T},\overline{\mathbf{h}}\!\!~_{\mathbf{y}_i}^\mathcal{T})/\tau\big)}{\sum\nolimits_{\mathbf{y}_j\in\{\mathcal{Y}^\mathcal{T}\}}\exp\big(g(\mathbf{h}_i^\mathcal{T}, \overline{\mathbf{h}}\!\!~_{\mathbf{y}_j}^\mathcal{T})/\tau\big)}\right],\!\!
\label{eq:finetune_loss}
\end{equation}
where $\overline{\mathbf{h}}\!\!~_{\mathbf{y}_i}^\mathcal{T}$ is the class prototype for samples in class $\mathbf{y}_i$.
We analyse the fine-tuning pipeline in Algorithm~\ref{alg:finetune} with complexity analysis.

\subsection{Algorithms and Complexity}
Shown in Appendix~\ref{app:alg}, \modelname~consists of two stages.
In pre-training, dual-view encoding and self-supervised alignment dominate the cost, yielding $\mathcal{O}(L(E_{\mathcal B}+|\mathcal B|)d+|\mathcal B|^2 d)$ per iteration, where $E_{\mathcal B}$ is the edge count in batch $\mathcal B$. In fine-tuning, semantic retrieval and structural motif retrieval are combined via domain-gated fusion and prompt-based classification, giving $\mathcal{O}(m[\log M_{\text{text}}+n\log M_{\text{struct}}+(k+n+C)d])$ per iteration, with $M_{\text{text}}$ and $M_{\text{struct}}$ the database sizes and $C$ the class number. Retrieval adds only logarithmic overhead, while adaptation updates prompts instead of full parameters, ensuring much lower cost than end-to-end fine-tuning.

Overall, the complexity remains comparable to state-of-the-art GFMs while achieving superior efficiency in few-shot adaptation.

%% file: fig/framework_marked.tex
\begin{tikzpicture}[very thick, black]
\node[draw=none,fill=none] (image) at (0,0){\includegraphics{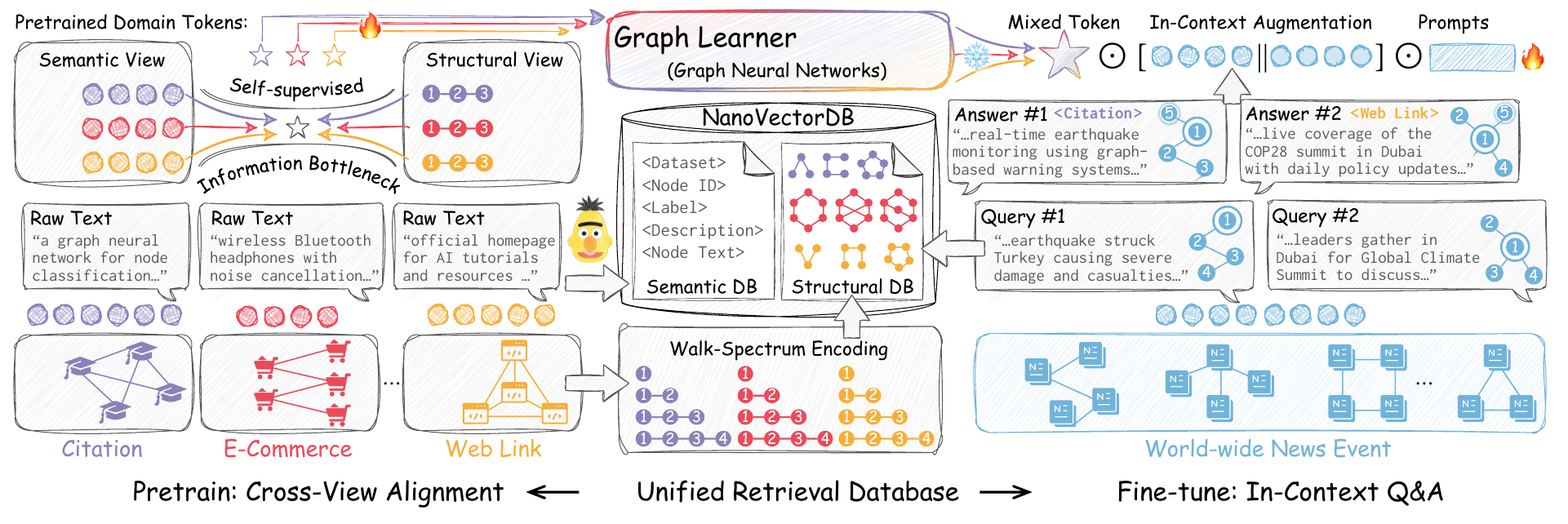}};


\node[align=center, color=black, font=\Huge] at (-15.29, -2.1) {$G_1^\mathcal{S}$};
\node[align=center, color=black, font=\Huge] at (-11.45, -2.1) {$G_2^\mathcal{S}$};
\node[align=center, color=black, font=\Huge] at (-7.2, -2.1) {$G_n^\mathcal{S}$};

\node[align=center, color=black, font=\Huge] at (4.59, -2.1) {$G_1^\mathcal{T}$};
\node[align=center, color=black, font=\Huge] at (7.55, -2.1) {$G_2^\mathcal{T}$};
\node[align=center, color=black, font=\Huge] at (10.85, -2.1) {$G_3^\mathcal{T}$};
\node[align=center, color=black, font=\Huge] at (14, -2.1) {$G_m^\mathcal{T}$};

\node[align=center, color=black, font=\huge] at (15.2, 4.88) {$\boldsymbol{\mathcal{P}}_{\boldsymbol{\Omega}}$};

\node[align=center, color=black, font=\Huge] at (-15.18, 2.775) {$\mathbf{H}^\text{text}$};

\node[align=center, color=black, font=\Huge] at (-4.95, 2.775) {$\mathbf{H}^\text{struct}$};

\node[align=center, color=black, font=\Huge] at (1.91, 4.53) {$h=g(f_{\boldsymbol{\Theta}}(\cdot))$};

\node[align=center, color=black, font=\Huge] at (-3.5, -4.86) {\hyperref[sec:db]{\textbf{(1)}}};
\node[align=center, color=black, font=\Huge] at (-14, -4.86) {\hyperref[sec:align]{\textbf{(2)}}};
\node[align=center, color=black, font=\Huge] at (6.48, -4.86) {\hyperref[sec:qa]{\textbf{(3)}}};

\end{tikzpicture}

%% file: 5_experiment.tex
\section{Experiment}
\label{sec:exp}
We evaluate \modelname\footnote{\url{https://github.com/RingBDStack/RAG-GFM}.}, focusing on the these research questions:
\begin{itemize}[leftmargin=*]
    \item \textbf{\textit{RQ1:}} How effective on cross-dataset or cross-domain few-shot node and graph classification? ($\rhd$~Section~\ref{sec:rq1})
    \item \textbf{\textit{RQ2:}} Which module contributes most? ($\rhd$~Section~\ref{sec:rq2})
    \item \textbf{\textit{RQ3:}} Can LLM achieve zero-shot reasoning? ($\rhd$~Section~\ref{sec:rq3})
    \item \textbf{\textit{RQ4:}} How efficient in time and memory? ($\rhd$~Section~\ref{sec:rq4})
    \item \textbf{\textit{RQ5:}} How reliable and interpretable is RAG?  ($\rhd$~Section~\ref{sec:rq5})
    \item \textbf{\textit{RQ6:}}  How sensitive to hyper-parameter changes? ($\rhd$~Section~\ref{sec:rq6})
\end{itemize}

\subsection{Experimental Settings}
\label{sec:exp_setting}
\subsubsection{\textbf{Datasets}}
To emphasize the pre-training capability across heterogeneous domains, we adopt \textit{\textbf{five}} benchmark text-attributed graph datasets spanning \textit{\textbf{three}} distinct domains. This design contrasts with conventional settings that often regard a single dataset as an independent domain, offering a more challenging evaluation.
\begin{itemize}[leftmargin=*]
    \item \textbf{{Citation} Domain:}
    {\texttt{Cora}}~\cite{mccallum2000automating}, {\texttt{CiteSeer}}~\cite{giles1998citeseer}, {\texttt{PubMed}}~\cite{sen2008collective}.
    \item \textbf{E-Commerce Domain:} {\texttt{Ogbn-Products}}~\cite{hu2020open} from a large-scale product co-purchase network, which includes sub-categories.
    \item \textbf{Web Link Domain:} {\texttt{Wiki-CS}}~\cite{mernyei2020wiki}, a hyperlink web page network constructed from a subset of Wikipedia.
\end{itemize}

\subsubsection{\textbf{Baselines}}
We compare \modelname~with \textit{\textbf{13}} state-of-the-art baselines from \textit{\textbf{four}} primary categories.
\begin{itemize}[leftmargin=*]
    \item \textbf{Vanilla GNNs:} \texttt{GCN}~\cite{kipf2022semi} and \texttt{GAT}~\cite{velickovic2017graph} without pre-training.
    \item \textbf{Graph Pre-training:} \texttt{DGI}~\cite{veličković2018deep}, \texttt{InfoGraph}~\cite{Sun2020InfoGraph}, \texttt{GraphCL}~\cite{you2020graph}.
    \item \textbf{Text-free GFMs:} \texttt{GCOPE}~\cite{zhao2024all}, \texttt{MDGPT}~\cite{yu2024text}, \texttt{SAMGPT}~\cite{yu2025samgpt}, and \texttt{MDGFM} \cite{wang2025multi}, which are evaluated on text-free graphs.
    \item \textbf{Text-attributed GFMs:} \texttt{OFA}~\cite{liu2023one}, \texttt{ZeroG}~\cite{li2024zerog}, \texttt{GraphCLIP}~\cite{zhu2024graphclip}, and \texttt{UniGraph}~\cite{he2024unigraph}, which are evaluated on text-attributed graphs.
\end{itemize}

\subsubsection{\textbf{Pre-training and Fine-tuning Settings}}
We evaluate node- and graph-level classification under the $m$-shot setting, where $m$ labeled samples per class are randomly selected. For graph task, ego-graphs centered on target nodes are extracted and labeled by central nodes~\cite{lu2021learning,yu2024hgprompt,yu2024text}. To assess generalization, we adopt two leave-out strategies, both referred to as \textbf{LODO}: \textbf{(1)} \underline{\textbf{L}}eave-\underline{\textbf{O}}ne-\underline{\textbf{D}}ataset-\underline{\textbf{O}}ut, holding out one dataset as target; and \textbf{(2)} \underline{\textbf{L}}eave-\underline{\textbf{O}}ne-\underline{\textbf{D}}omain-\underline{\textbf{O}}ut, excluding an entire domain during pre-training. These variants capture transferability across unseen datasets and unseen domains. Results are reported by mean values with standard deviation.

\input{table/res_main}
\subsection{\textit{RQ1:} Transfer across Domains and Tasks}
\label{sec:rq1}
Table~\ref{tab:rq1} reports the results of few-shot node and graph classification under both \textbf{LODO} settings.
Results reveal that:

\textbf{(1) Overall superiority.}  
The proposed \modelname~consistently outperforms all baselines over each target graph. The advantage is most evident in the challenging \textbf{LODO (domain)} case, where it raises the 5-shot graph classification accuracy on \texttt{Wiki-CS} by over 5.3\% compared with \texttt{UniGraph}, relatively. Baselines generally struggle as they compress knowledge entirely into parameters or rely only on texts, limiting generalization to unseen domains.

\textbf{(2) Retrieval enhances transfer.}  
In \textbf{LODO (dataset)} setting, \modelname~consistently outperforms parameter-only GFMs, with the average relative gains of \textasciitilde 3.0\%. While baselines can still leverage shared domain priors, their parameter-centric representations fail to capture sufficient diversity across datasets. By contrast, retrieval from the unified database introduces complementary evidence: semantic queries supply textual signals, and structural queries provide transferable motifs, enabling adaptation with minimal supervision.  

\textbf{(3) Cross-view alignment strengthens cross-domain robustness.}  
In the stricter \textbf{LODO (domain)} setting, where the entire target domain is unseen, the performance gap widens further with an average relative improvement of \textasciitilde 4\%. Baselines relying on text-only or domain-specific features degrade sharply, since they cannot bridge modality and domain gaps. In contrast, cross-view alignment in \modelname~enforces consistency between semantic and structural views, reducing overfitting to pre-training domains and ensuring that retrieved knowledge remains useful. 

\textbf{(4) Domain-gated prompting ensures universality.}  
Consistent gains across tasks (on average 4.5\% higher accuracy in the node task and 3.8\% in the graph task, relatively) demonstrate that the framework is not tailored to a specific scenario. Baselines often overfit to one task formulation: models tuned for node classification transfer less effectively to graph classification. By introducing domain-gated prompts, our \modelname~adapts flexibly to both granularities, which is particularly advantageous in few-shot scenarios where labeled data is extremely scarce.

\begin{figure*}[!t]
    \begin{minipage}{0.33\textwidth}
        \centering
        \includegraphics[width=\textwidth]{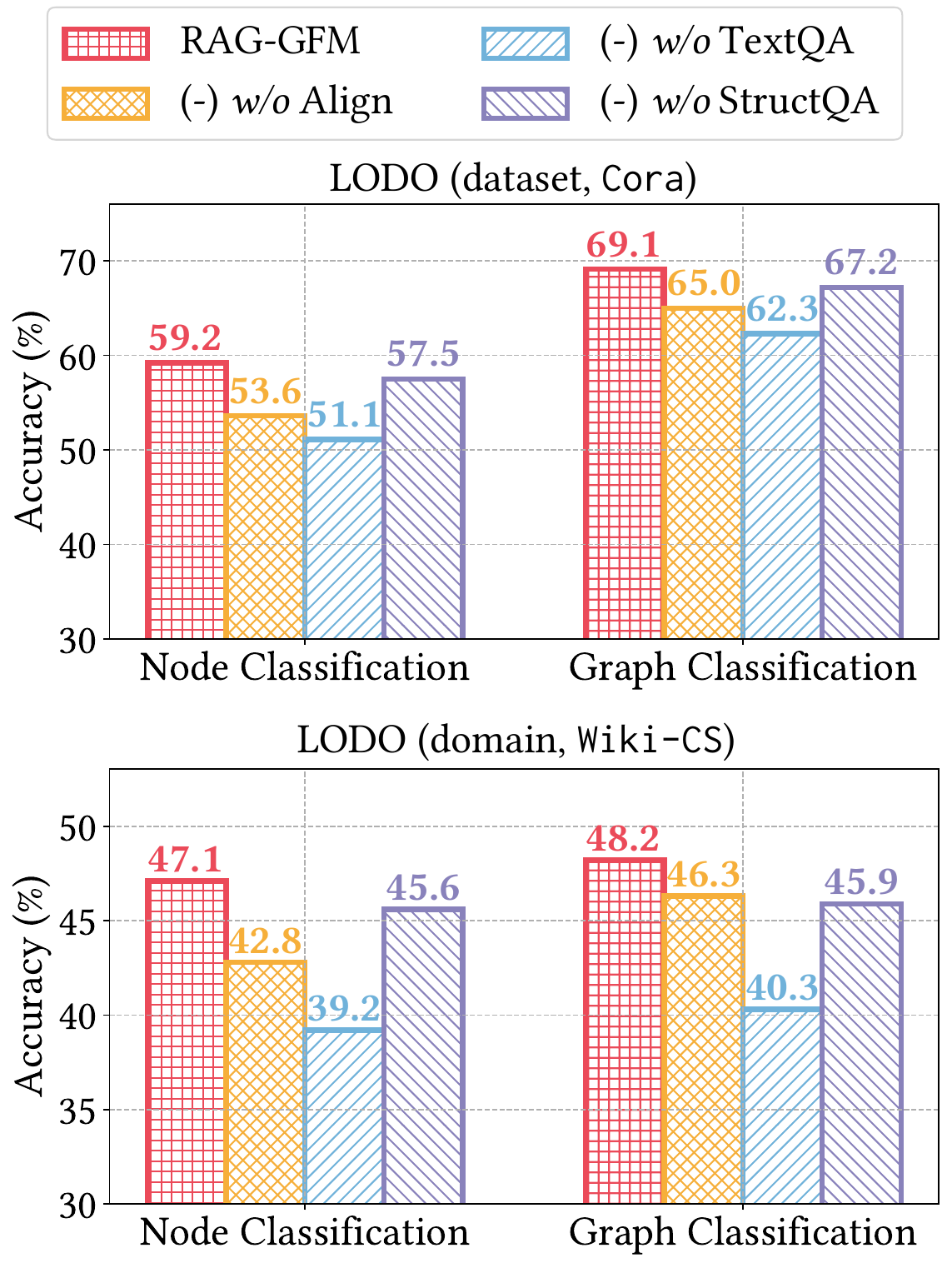}
        \vspace{-0.7cm}
        \caption{Ablation Study.}
        \label{fig:abla}
    \end{minipage} 
    \begin{minipage}{0.33\textwidth}
        \centering
        \includegraphics[width=\textwidth]{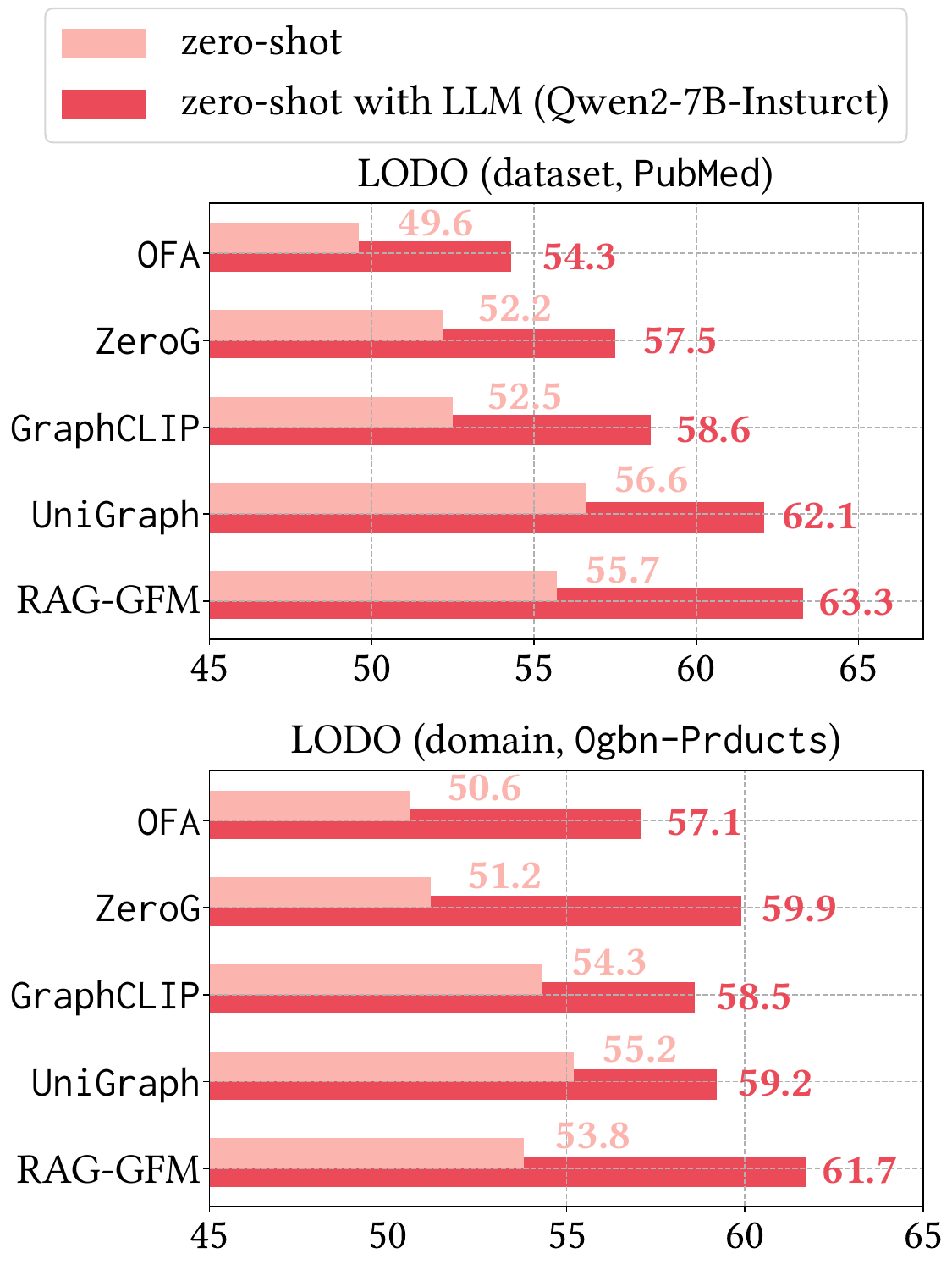}
        \vspace{-0.7cm}
        \caption{Zero-shot Reasoning.}
        \label{fig:llm}
    \end{minipage}
    \begin{minipage}{0.33\textwidth}
        \centering
        \includegraphics[width=\textwidth]{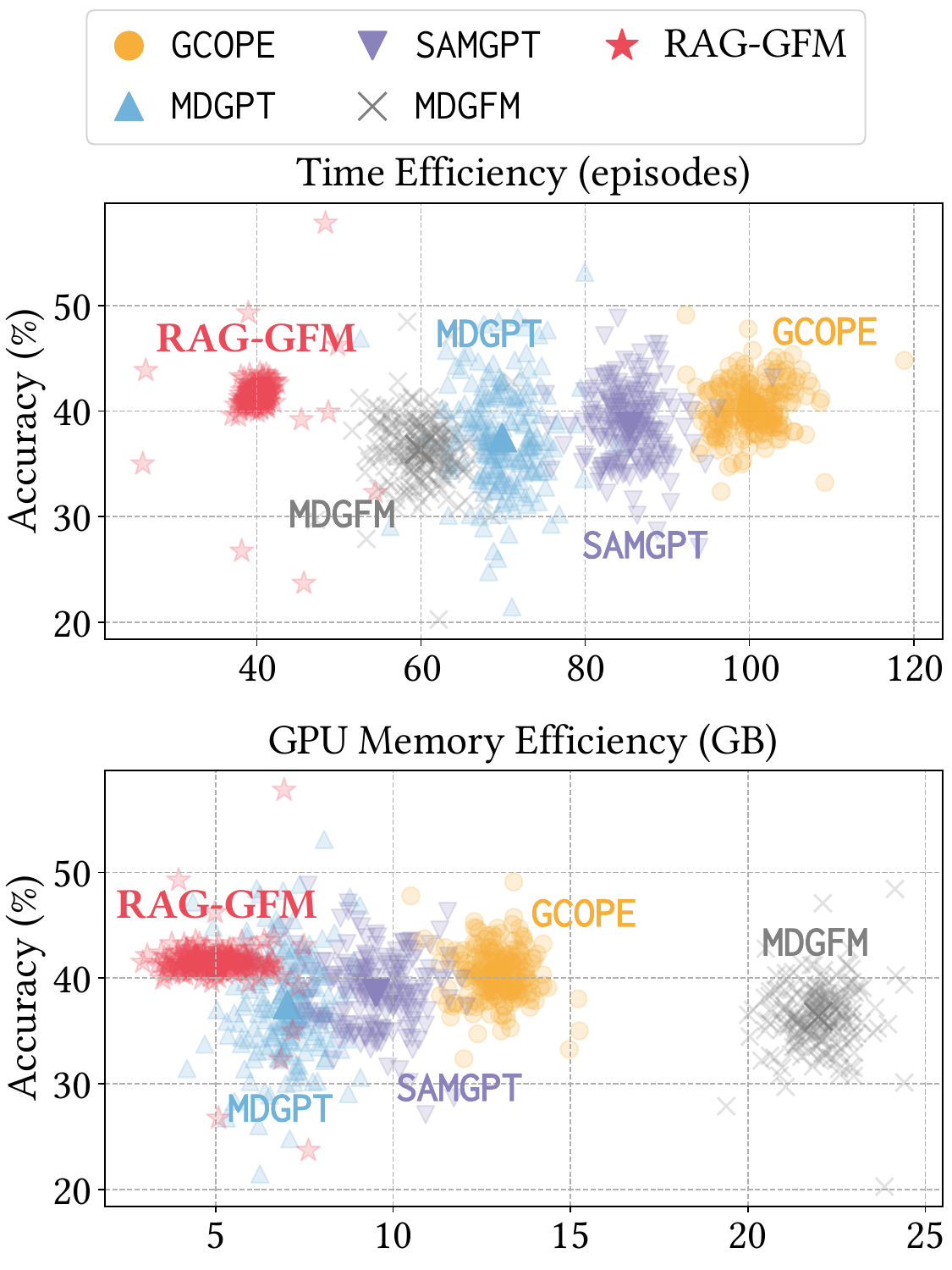}
        \vspace{-0.7cm}
        \caption{Efficiency Analysis on \texttt{CiteSeer}.}
        \label{fig:efficiency}
    \end{minipage}
\end{figure*}

\subsection{\textit{RQ2:} Ablation Study}
\label{sec:rq2}
We conduct ablation studies on \textit{\textbf{three}} core modules:
\begin{itemize}[leftmargin=*]
    \item \textbf{\modelname~(\textit{w/o} Align)}: remove the cross-view knowledge alignment in pre-training (Section~\ref{sec:align}). Semantic and structural encoders are trained independently without mutual consistency.
    \item \textbf{\modelname~(\textit{w/o} TextQA)}: remove the semantic retrieval in fine-tuning (Section~\ref{sec:qa}). Textual augmentation is disabled and relies only on structural retrieval and parameterized features.
    \item \textbf{\modelname~(\textit{w/o} StructQA)}: remove the structural retrieval in fine-tuning (Section~\ref{sec:qa}). Structural augmentation is discarded, leaving only textual retrieval and parameterized features.
\end{itemize}

Results in Figure~\ref{fig:abla} demonstrate the full \modelname~achieves the best results across both settings. \modelname~(\textit{w/o} Align) causes clear drops (\eg, 59.2\% to 53.6\% on \texttt{Cora}), underscoring the importance of semantic-structural consistency in pre-training. \modelname~(\textit{w/o} TextQA) leads to the largest decline (nearly 8\% on \texttt{Wiki-CS}), showing that raw attributes alone are insufficient and external semantic evidence is essential. \modelname~(\textit{w/o} StructQA) also reduces accuracy (\eg, 69.1\% to 67.2\% on \texttt{Cora}), though less severely, indicating that motif-level cues provide secondary but stable benefits.


\subsection{\textit{RQ3:} Zero-shot Reasoning with LLMs}
\label{sec:rq3}
To in-depth examine the potential of large language models (LLMs), we evaluate a \textit{\textbf{zero-shot}} setting without fine-tuning. Two scenarios are compared: \textbf{(1) zero-shot}, where the pre-trained models directly predict without supervision, and \textbf{(2) zero-shot with LLM}, where the graph task is reformulated into language queries (\eg, \textit{``Which class does this node belong to?''}). Each target node is augmented with retrieved textual and structural context from the pre-training database, concatenated with its raw description, and fed into an LLM (we use Qwen2-7B-Instruct~\cite{team2024qwen2}) to produce predictions. This setup allows us to assess whether external language priors can compensate for the absence of labeled examples.

As baselines, we select GFMs for text-attributed graphs, most of which already leverage LLMs as feature enhancers or semantic aligners during pre-training. However, these designs do not directly test whether LLMs themselves can serve as zero-shot reasoners.

Results in Figure~\ref{fig:llm} demonstrate that while \modelname~is competitive as an LLM-free GFM, it is sometimes slightly behind LLM-enhanced baselines in the zero-shot case. Notably, once equipped with LLM reasoning, it consistently achieves the best performance, improving from 55.7\% to 63.3\% on \texttt{PubMed} and from 53.8\% to 61.7\% on \texttt{Ogbn-Products}, surpassing all baselines. More importantly, the gains are not simply due to invoking stronger LLMs: by grounding reasoning in our unified dual-modal retrieval database, the prompts provide structured, domain-aligned evidence that enables LLMs to generalize more faithfully across unseen graphs. Furthermore, even existing LLM-enhanced GFMs benefit from our retrieval-augmented prompting, highlighting that \modelname~is not only effective in its own design but also serves as a general, pluggable enhancement that can universally elevate zero-shot reasoning in graph learning.

\subsection{\textit{RQ4:} Time and Memory Efficiency}
\label{sec:rq4}
We further compare \modelname~with four state-of-the-art text-free GFMs in terms of fine-tuning efficiency on \texttt{CiteSeer} under \textbf{LODO (dataset)}. We report both the number of episodes required to reach stable accuracy and the peak GPU memory usage. As shown in Figure~\ref{fig:efficiency}, \modelname~achieves clear advantages on both fronts. In terms of time, it converges much faster, as retrieval-augmented prompts inject external knowledge directly without costly parameter updates. For memory, most knowledge is externalized to the dual-modal database, where only lightweight prompts are optimized while encoders remain frozen, reducing GPU usage to less than half of \texttt{MDGFM}. Although \texttt{GCOPE} and \texttt{SAMGPT} can reach comparable accuracy, they require 2-3$\times$ more episodes and substantially higher memory, which limits their scalability in practice.

\begin{figure*}[!t]
    \begin{minipage}{0.33\textwidth}
        \centering
        \includegraphics[width=\textwidth]{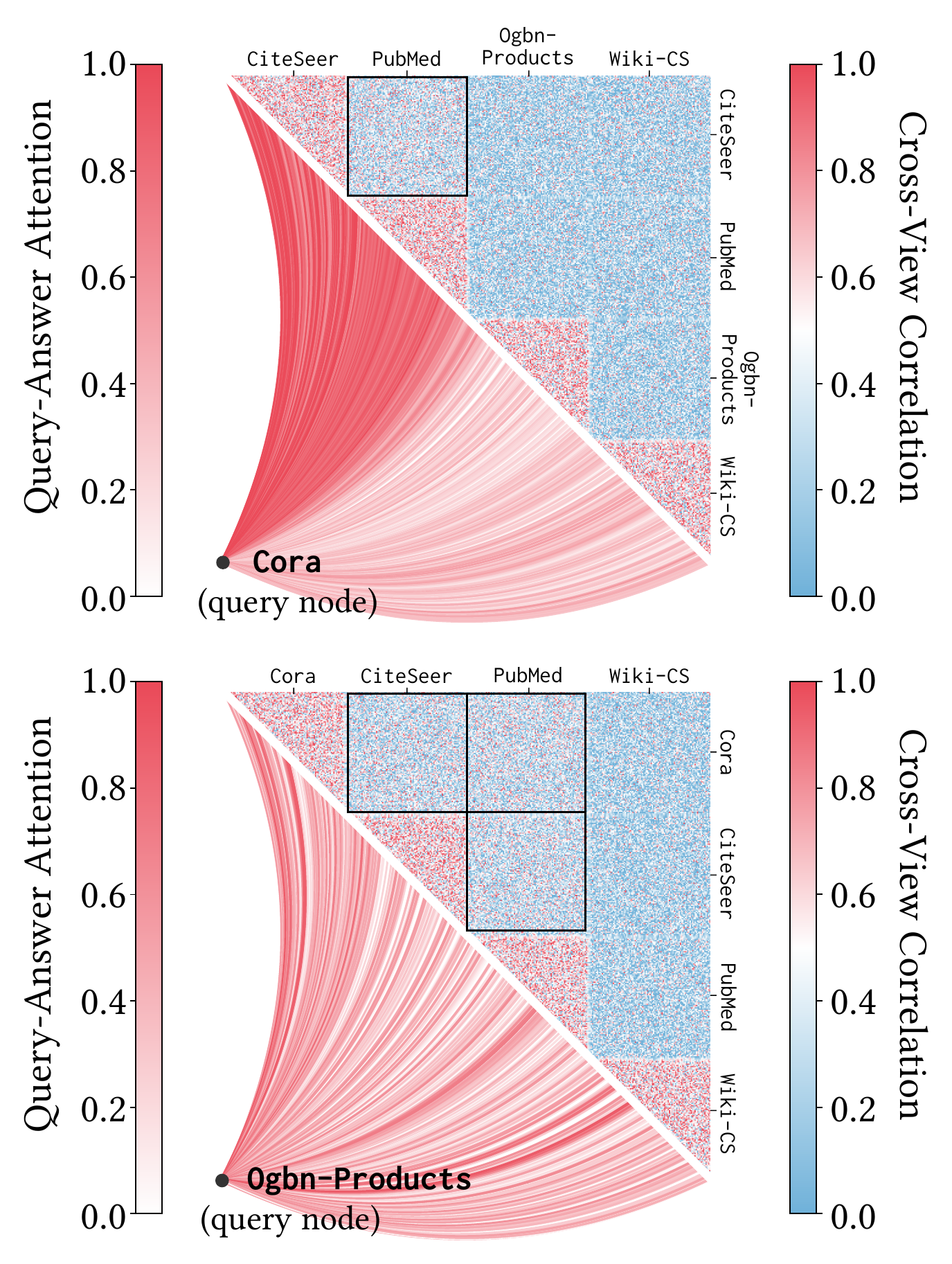}
        \vspace{-0.7cm}
        \caption{RAG Correlation Map.}
        \label{fig:heat}
    \end{minipage} 
    \begin{minipage}{0.33\textwidth}
        \centering
        \includegraphics[width=\textwidth]{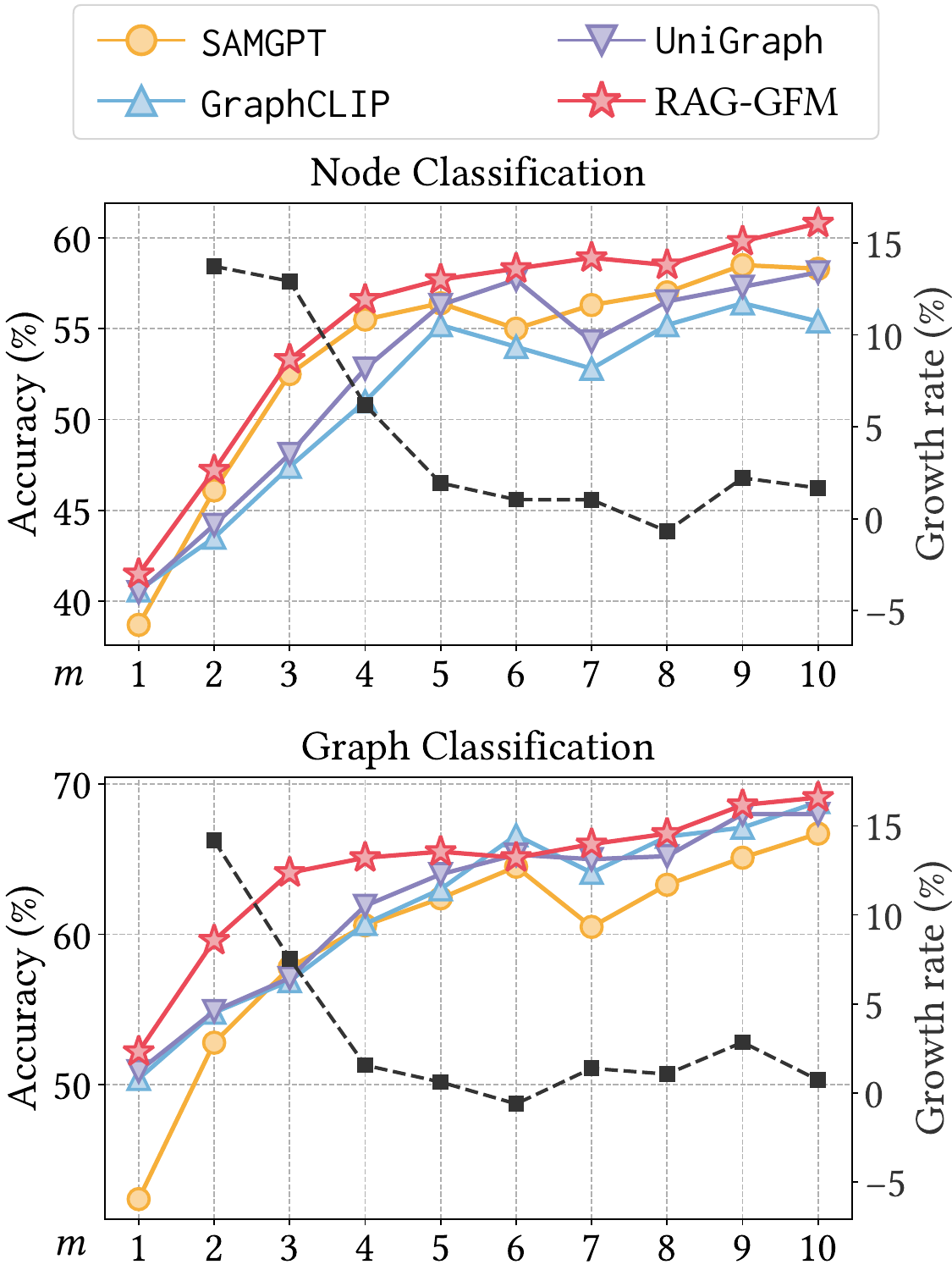}
        \vspace{-0.7cm}
        \caption{$\boldsymbol{m}$-Shot Classification (\texttt{CiteSeer}).}
        \label{fig:mshot}
    \end{minipage}
    \begin{minipage}{0.33\textwidth}
        \centering
        \includegraphics[width=\textwidth]{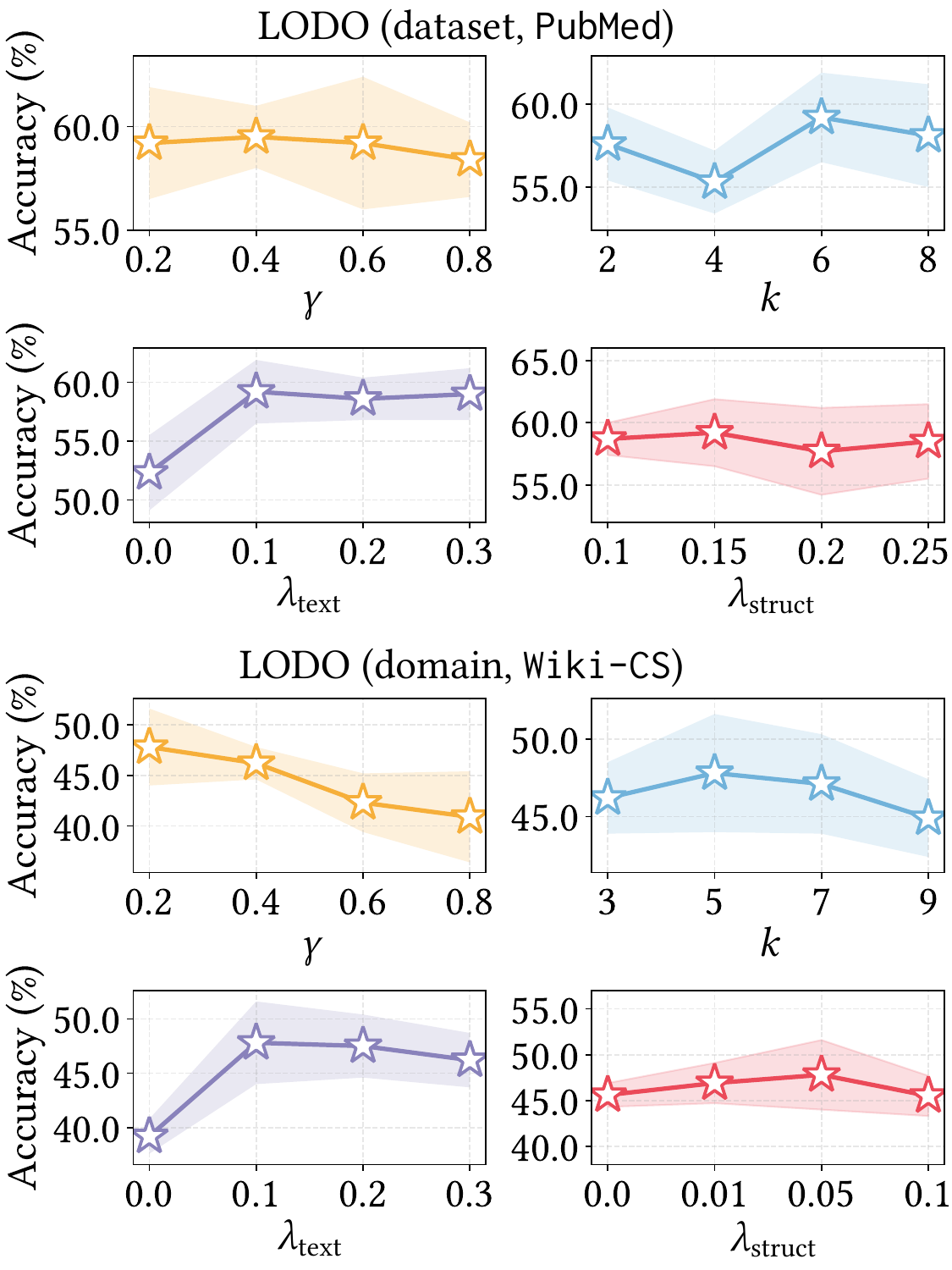}
        \vspace{-0.7cm}
        \caption{Hyper-parameter Analysis.}
        \label{fig:sensitivity}
    \end{minipage}
\end{figure*}

\subsection{\textit{RQ5:} Reliability and Interpretability of RAG}
\label{sec:rq5}

We assess the reliability and interpretability of RAG by visualizing its retrieval behavior under \textbf{LODO (dataset)} and \textbf{LODO (domain)} settings. We construct a cross-view correlation map and query-answer attention visualization, where heatmaps capture semantic-structural correlations across source datasets, and curved lines indicate the attention between a query node and retrieved source nodes, with color and thickness reflecting attention intensity.

As shown in Figure~\ref{fig:heat}, clear diagonal blocks emerge in the correlation maps, indicating strong semantic-structural consistency within datasets, while cross-dataset or cross-domain correlations remain low. Datasets from the same domain (\eg, \texttt{CiteSeer} and \texttt{PubMed}) still exhibit relatively higher correlations, suggesting transferable within-domain relations. In \textbf{LODO (dataset)}, a query from \texttt{Cora} assigns higher attention to citation-domain sources, reflecting adaptive retrieval of aligned knowledge. In \textbf{LODO (domain)}, attention becomes more evenly distributed across unseen domains while maintaining weak but informative focus on partially aligned datasets such as \texttt{Cora} and \texttt{Wiki-CS}.

\subsection{\textit{RQ6:} Sensitivity to Hyper-parameters}
\label{sec:rq6}
We evaluate the robustness of \modelname~under two groups.

\textbf{$\boldsymbol{m}$-Shot Classification.}
Figure~\ref{fig:mshot} presents the performance trends as $m$ increases of both node and graph classification tasks on the \textbf{LODO (dataset,} \texttt{CiteSeer}\textbf{)} setting.
All methods exhibit a saturation curve, where accuracy improves rapidly when moving from extremely low-shot (1-3 samples) to moderate-shot (5-6 samples) and then stabilizes.
Notably, \modelname~consistently outperforms all baselines at most of the shots, achieving higher accuracy and smoother convergence.
The black dashed line depicts its growth rate, showing a sharp improvement at early stages followed by stable gains, indicating that retrieval-augmented prompting accelerates label efficiency and mitigates overfitting in low-shot regimes.

\textbf{Sensitivity Analysis.}
Results are shown in Figure~\ref{fig:sensitivity}. Across both \textbf{LODO (dataset,} \texttt{PubMed}\textbf{)} and \textbf{LODO (domain,} \texttt{Wiki-CS}\textbf{)} settings, performance remains stable under moderate perturbations. For $\gamma$, the weight of the domain-token regularizer in Eq.~\eqref{eq:pretrain_loss}, overly large values degrade performance by suppressing domain priors. For $k$, the number of retrieved query-answer pairs (Section~\ref{sec:qanda}), moderate values best balance retrieval diversity and noise. Accuracy peaks near $\lambda_{\text{text}}=0.1$ in Eq.~\eqref{eq:augment_1}, indicating its benefits from moderate textual retrieval, while larger values cause drift. Similarly, moderate $\lambda_{\text{struct}}$ in Eq.~\eqref{eq:augment_2} yields stable performance, as excessive structural signals may introduce bias. Overall, these trends demonstrate robustness without fine-grained tuning.

%% file: table/res_main.tex
\begin{table*}[!t]
\renewcommand{\arraystretch}{0.85}
  \centering
  \caption{Few-shot classification results under the LODO setting. We report mean accuracy (\%) with standard deviation. ``LODO (dataset)'' denotes training on all datasets except the target, irrespective of domain. ``LODO (domain)'' denotes training with all datasets excluding those belonging to the target domain. Best results are presented in \textbf{bold} and the runner-ups are \underline{underlined}.}
  \resizebox{\textwidth}{!}{
    \begin{tabular}{lc>{\columncolor{black!6}}cc>{\columncolor{black!6}}cc>{\columncolor{black!6}}cc>{\columncolor{black!6}}cc>{\columncolor{black!6}}c}
    \toprule
    \textbf{Setting} & \multicolumn{6}{c}{\textbf{LODO (dataset)}}  & \multicolumn{4}{c}{\textbf{LODO (domain)}} \\
    \cmidrule{1-1} \cmidrule(l{1.5mm}r{0mm}){2-7} \cmidrule(l){8-11}
    \textbf{Target Dataset} & \multicolumn{2}{c}{\texttt{\textbf{Cora}}} & \multicolumn{2}{c}{\texttt{\textbf{CiteSeer}}} & \multicolumn{2}{c}{\texttt{\textbf{PubMed}}} & \multicolumn{2}{c}{\texttt{\textbf{Ogbn-Products}}} & \multicolumn{2}{c}{\textbf{\texttt{Wiki-CS}}} \\
    \cmidrule{1-1} \cmidrule(l{1.5mm}r{0mm}){2-7} \cmidrule(l){8-11}
    $\boldsymbol{m}$\textbf{-shot} & \textbf{1}     & \textbf{5}     & \textbf{1}     & \textbf{5}     & \textbf{1}     & \textbf{5}     & \textbf{1}     & \textbf{5}     & \textbf{1}     & \textbf{5} \\
    \midrule
    \textbf{Method} & \multicolumn{10}{c}{\textbf{Node Classification}} \\
    \midrule
    \texttt{GCN}~\scalebox{0.75}{~(\mycite{kipf2022semi}\textit{ICLR'17})}   & 28.4\scalebox{0.8}{\,±\,4.6} & 50.2\scalebox{0.8}{\,±\,4.9} & 29.3\scalebox{0.8}{\,±\,3.4} & 45.9\scalebox{0.8}{\,±\,5.4} & 40.3\scalebox{0.8}{\,±\,6.9} & 50.7\scalebox{0.8}{\,±\,7.5} & 44.7\scalebox{0.8}{\,±\,4.3} & 48.1\scalebox{0.8}{\,±\,3.4} & 37.2\scalebox{0.8}{\,±\,5.1} & 48.1\scalebox{0.8}{\,±\,4.9} \\
    \texttt{GAT}~\scalebox{0.75}{~(\mycite{velickovic2017graph}\textit{ICLR'18})}   & 29.7\scalebox{0.8}{\,±\,5.2} & 49.0\scalebox{0.8}{\,±\,7.9} & 29.3\scalebox{0.8}{\,±\,3.5} & 46.1\scalebox{0.8}{\,±\,5.1} & 40.5\scalebox{0.8}{\,±\,4.0} & 52.2\scalebox{0.8}{\,±\,6.3} & 44.6\scalebox{0.8}{\,±\,4.0} & 48.1\scalebox{0.8}{\,±\,4.5} & 37.9\scalebox{0.8}{\,±\,4.5} & 48.6\scalebox{0.8}{\,±\,4.5} \\
    \cmidrule{1-1} \cmidrule(l{1.5mm}r{0mm}){2-7} \cmidrule(l){8-11}
    \texttt{DGI}~\scalebox{0.75}{~(\mycite{veličković2018deep}\textit{ICLR'19})}   & 30.8\scalebox{0.8}{\,±\,3.9} & 49.9\scalebox{0.8}{\,±\,6.6} & 31.4\scalebox{0.8}{\,±\,4.1} & 46.5\scalebox{0.8}{\,±\,7.1} & 40.0\scalebox{0.8}{\,±\,5.9} & 53.6\scalebox{0.8}{\,±\,7.1} & 46.0\scalebox{0.8}{\,±\,5.4} & 50.1\scalebox{0.8}{\,±\,4.2} & 38.1\scalebox{0.8}{\,±\,5.1} & 49.2\scalebox{0.8}{\,±\,4.4} \\
    \texttt{GraphCL}~\scalebox{0.75}{~(\mycite{you2020graph}\textit{NeurIPS'20})} & 33.6\scalebox{0.8}{\,±\,5.8} & 53.2\scalebox{0.8}{\,±\,5.4} & 28.2\scalebox{0.8}{\,±\,3.1} & 48.8\scalebox{0.8}{\,±\,7.7} & 39.0\scalebox{0.8}{\,±\,8.7} & 54.7\scalebox{0.8}{\,±\,4.4} & 46.1\scalebox{0.8}{\,±\,5.0} & 50.5\scalebox{0.8}{\,±\,4.6} & 40.0\scalebox{0.8}{\,±\,4.0} & 50.1\scalebox{0.8}{\,±\,5.2} \\
    \cmidrule{1-1} \cmidrule(l{1.5mm}r{0mm}){2-7} \cmidrule(l){8-11}
    \texttt{GCOPE}~\scalebox{0.75}{~(\mycite{zhao2024all}\textit{KDD'24})} & 36.3\scalebox{0.8}{\,±\,3.9} & 55.6\scalebox{0.8}{\,±\,6.4} & 40.4\scalebox{0.8}{\,±\,4.6} & 56.9\scalebox{0.8}{\,±\,5.8} & 44.8\scalebox{0.8}{\,±\,4.7} & 53.6\scalebox{0.8}{\,±\,8.6} & 47.7\scalebox{0.8}{\,±\,4.9} & 51.4\scalebox{0.8}{\,±\,3.3} & 45.8\scalebox{0.8}{\,±\,5.5} & 53.5\scalebox{0.8}{\,±\,4.7} \\
    \texttt{MDGPT}~\scalebox{0.75}{~(\mycite{yu2024text}\textit{arXiv'24})} & 42.6\scalebox{0.8}{\,±\,6.8} & 62.7\scalebox{0.8}{\,±\,6.0} & 37.9\scalebox{0.8}{\,±\,7.2} & 55.9\scalebox{0.8}{\,±\,3.3} & 51.0\scalebox{0.8}{\,±\,9.0} & 58.7\scalebox{0.8}{\,±\,6.2} & 49.1\scalebox{0.8}{\,±\,6.0} & 56.6\scalebox{0.8}{\,±\,2.7} & 45.0\scalebox{0.8}{\,±\,4.8} & 54.1\scalebox{0.8}{\,±\,5.2} \\
    \texttt{SAMGPT}~\scalebox{0.75}{~(\mycite{yu2025samgpt}\textit{WWW'25})} & 46.8\scalebox{0.8}{\,±\,6.5} & 64.6\scalebox{0.8}{\,±\,6.7} & 38.7\scalebox{0.8}{\,±\,6.4} & 56.4\scalebox{0.8}{\,±\,4.7} & 51.9\scalebox{0.8}{\,±\,9.5} & 59.1\scalebox{0.8}{\,±\,6.0} & 49.8\scalebox{0.8}{\,±\,4.4} & 56.2\scalebox{0.8}{\,±\,3.3} & 44.4\scalebox{0.8}{\,±\,5.5} & 54.4\scalebox{0.8}{\,±\,5.8} \\
    \texttt{MDGFM}~\scalebox{0.75}{~(\mycite{wang2025multi}\textit{ICML'25})} & 47.4\scalebox{0.8}{\,±\,6.3} & 66.0\scalebox{0.8}{\,±\,6.5} & 36.3\scalebox{0.8}{\,±\,6.2} & 55.8\scalebox{0.8}{\,±\,4.0} & 50.2\scalebox{0.8}{\,±\,8.8} & 58.4\scalebox{0.8}{\,±\,6.4} & 48.5\scalebox{0.8}{\,±\,4.7} & 54.7\scalebox{0.8}{\,±\,4.9} & 43.4\scalebox{0.8}{\,±\,5.8} & 53.9\scalebox{0.8}{\,±\,4.4} \\
    \cmidrule{1-1} \cmidrule(l{1.5mm}r{0mm}){2-7} \cmidrule(l){8-11}
    \texttt{OFA}~\scalebox{0.75}{~(\mycite{liu2023one}\textit{ICLR'24})}   & 45.9\scalebox{0.8}{\,±\,6.3} & 67.7\scalebox{0.8}{\,±\,2.9} & 38.0\scalebox{0.8}{\,±\,7.6} & 52.8\scalebox{0.8}{\,±\,6.4} & 46.3\scalebox{0.8}{\,±\,6.0} & 56.0\scalebox{0.8}{\,±\,5.9} & 49.1\scalebox{0.8}{\,±\,5.7} & 55.3\scalebox{0.8}{\,±\,4.2} & 42.8\scalebox{0.8}{\,±\,4.6} & 54.3\scalebox{0.8}{\,±\,4.0} \\
    \texttt{ZeroG}~\scalebox{0.75}{~(\mycite{li2024zerog}\textit{KDD'24})} & 51.8\scalebox{0.8}{\,±\,5.6} & 71.4\scalebox{0.8}{\,±\,1.7} & 39.7\scalebox{0.8}{\,±\,5.9} & 54.6\scalebox{0.8}{\,±\,2.0} & 53.1\scalebox{0.8}{\,±\,3.5} & 63.0\scalebox{0.8}{\,±\,3.5} & 53.2\scalebox{0.8}{\,±\,2.9} & 59.9\scalebox{0.8}{\,±\,3.1} & \underline{46.1\scalebox{0.8}{\,±\,3.4}} & 59.0\scalebox{0.8}{\,±\,2.0} \\
    \texttt{GraphCLIP}~\scalebox{0.75}{~(\mycite{zhu2024graphclip}\textit{WWW'25})} & 53.9\scalebox{0.8}{\,±\,5.3} & 73.1\scalebox{0.8}{\,±\,2.9} & \underline{40.6\scalebox{0.8}{\,±\,3.4}} & 55.2\scalebox{0.8}{\,±\,1.2} & 56.8\scalebox{0.8}{\,±\,1.9} & 65.2\scalebox{0.8}{\,±\,2.8} & 53.4\scalebox{0.8}{\,±\,6.1} & \underline{62.6\scalebox{0.8}{\,±\,4.3}} & 45.5\scalebox{0.8}{\,±\,2.1} & \underline{59.9\scalebox{0.8}{\,±\,2.8}} \\
    \texttt{UniGraph}~\scalebox{0.75}{~(\mycite{he2024unigraph}\textit{KDD'25})} & \underline{56.1\scalebox{0.8}{\,±\,6.3}} & \underline{74.8\scalebox{0.8}{\,±\,1.9}} & 40.5\scalebox{0.8}{\,±\,3.1} & \underline{56.3\scalebox{0.8}{\,±\,2.2}} & \underline{57.0\scalebox{0.8}{\,±\,3.1}} & \underline{66.8\scalebox{0.8}{\,±\,2.4}} & \underline{53.8\scalebox{0.8}{\,±\,3.4}} & 61.0\scalebox{0.8}{\,±\,3.5} & 45.1\scalebox{0.8}{\,±\,3.6} & 58.4\scalebox{0.8}{\,±\,3.1} \\
    \cmidrule{1-1} \cmidrule(l{1.5mm}r{0mm}){2-7} \cmidrule(l){8-11}
    \textbf{\modelname~\textit{(ours)}} & \textbf{58.4\scalebox{0.8}{\,±\,6.0}} & \textbf{76.1\scalebox{0.8}{\,±\,0.7}} & \textbf{41.5\scalebox{0.8}{\,±\,3.0}} & \textbf{57.7\scalebox{0.8}{\,±\,1.8}} & \textbf{59.2\scalebox{0.8}{\,±\,2.7}} & \textbf{68.7\scalebox{0.8}{\,±\,1.8}} & \textbf{55.4\scalebox{0.8}{\,±\,7.6}} & \textbf{64.2\scalebox{0.8}{\,±\,4.4}} & \textbf{47.8\scalebox{0.8}{\,±\,3.8}} & \textbf{60.9\scalebox{0.8}{\,±\,2.5}} \\
    \midrule[0.7pt]
    \textbf{Method} & \multicolumn{10}{c}{\textbf{Graph Classification}} \\
    \midrule
    \texttt{GCN}~\scalebox{0.75}{~(\mycite{kipf2022semi}\textit{ICLR'17})}   & 40.1\scalebox{0.8}{\,±\,4.8} & 52.9\scalebox{0.8}{\,±\,4.1} & 29.5\scalebox{0.8}{\,±\,5.7} & 43.9\scalebox{0.8}{\,±\,5.9} & 45.3\scalebox{0.8}{\,±\,7.3} & 55.4\scalebox{0.8}{\,±\,5.3} & 47.6\scalebox{0.8}{\,±\,3.2} & 52.6\scalebox{0.8}{\,±\,5.3} & 38.9\scalebox{0.8}{\,±\,4.1} & 41.5\scalebox{0.8}{\,±\,6.4} \\
    \texttt{GAT}~\scalebox{0.75}{~(\mycite{velickovic2017graph}\textit{ICLR'18})}   & 36.0\scalebox{0.8}{\,±\,5.1} & 49.6\scalebox{0.8}{\,±\,5.1} & 26.0\scalebox{0.8}{\,±\,7.8} & 45.3\scalebox{0.8}{\,±\,7.3} & 41.0\scalebox{0.8}{\,±\,5.8} & 54.5\scalebox{0.8}{\,±\,7.3} & 49.2\scalebox{0.8}{\,±\,5.6} & 52.9\scalebox{0.8}{\,±\,5.5} & 38.3\scalebox{0.8}{\,±\,4.6} & 41.1\scalebox{0.8}{\,±\,3.5} \\
    \cmidrule{1-1} \cmidrule(l{1.5mm}r{0mm}){2-7} \cmidrule(l){8-11}
    \texttt{InfoGraph}~\scalebox{0.75}{~(\mycite{Sun2020InfoGraph}\textit{ICLR'20})} & 42.2\scalebox{0.8}{\,±\,5.2} & 54.7\scalebox{0.8}{\,±\,4.9} & 30.2\scalebox{0.8}{\,±\,4.1} & 47.2\scalebox{0.8}{\,±\,5.2} & 49.1\scalebox{0.8}{\,±\,5.4} & 59.7\scalebox{0.8}{\,±\,7.1} & 50.7\scalebox{0.8}{\,±\,4.3} & 53.8\scalebox{0.8}{\,±\,5.1} & 40.4\scalebox{0.8}{\,±\,4.3} & 42.4\scalebox{0.8}{\,±\,5.0} \\
    \texttt{GraphCL}~\scalebox{0.75}{~(\mycite{you2020graph}\textit{NeurIPS'20})} & 39.6\scalebox{0.8}{\,±\,5.8} & 55.2\scalebox{0.8}{\,±\,5.9} & 32.6\scalebox{0.8}{\,±\,6.5} & 46.4\scalebox{0.8}{\,±\,3.8} & 47.7\scalebox{0.8}{\,±\,7.0} & 60.0\scalebox{0.8}{\,±\,5.4} & 51.7\scalebox{0.8}{\,±\,5.9} & 53.0\scalebox{0.8}{\,±\,5.2} & 40.8\scalebox{0.8}{\,±\,4.6} & 42.5\scalebox{0.8}{\,±\,4.3} \\
    \cmidrule{1-1} \cmidrule(l{1.5mm}r{0mm}){2-7} \cmidrule(l){8-11}
    \texttt{GCOPE}~\scalebox{0.75}{~(\mycite{zhao2024all}\textit{KDD'24})} & 55.9\scalebox{0.8}{\,±\,7.4} & 63.9\scalebox{0.8}{\,±\,4.8} & 41.0\scalebox{0.8}{\,±\,9.0} & 58.2\scalebox{0.8}{\,±\,5.8} & 54.4\scalebox{0.8}{\,±\,8.6} & 66.4\scalebox{0.8}{\,±\,3.7} & 55.8\scalebox{0.8}{\,±\,4.3} & 57.7\scalebox{0.8}{\,±\,4.8} & 42.2\scalebox{0.8}{\,±\,5.8} & 49.8\scalebox{0.8}{\,±\,3.5} \\
    \texttt{MDGPT}~\scalebox{0.75}{~(\mycite{yu2024text}\textit{arXiv'24})} & 52.8\scalebox{0.8}{\,±\,6.7} & 65.1\scalebox{0.8}{\,±\,4.2} & 41.0\scalebox{0.8}{\,±\,9.7} & 59.3\scalebox{0.8}{\,±\,6.0} & 55.5\scalebox{0.8}{\,±\,8.3} & 67.6\scalebox{0.8}{\,±\,4.6} & 54.5\scalebox{0.8}{\,±\,4.8} & 60.5\scalebox{0.8}{\,±\,3.4} & 43.2\scalebox{0.8}{\,±\,6.2} & 48.9\scalebox{0.8}{\,±\,4.2} \\
    \texttt{SAMGPT}~\scalebox{0.75}{~(\mycite{yu2025samgpt}\textit{WWW'25})} & 53.3\scalebox{0.8}{\,±\,4.3} & 69.3\scalebox{0.8}{\,±\,3.4} & 42.4\scalebox{0.8}{\,±\,7.3} & 62.4\scalebox{0.8}{\,±\,5.7} & 57.7\scalebox{0.8}{\,±\,6.3} & 68.0\scalebox{0.8}{\,±\,4.6} & 54.4\scalebox{0.8}{\,±\,3.2} & 60.8\scalebox{0.8}{\,±\,4.8} & 43.5\scalebox{0.8}{\,±\,5.6} & 48.3\scalebox{0.8}{\,±\,5.7} \\
    \texttt{MDGFM}~\scalebox{0.75}{~(\mycite{wang2025multi}\textit{ICML'25})} & 55.5\scalebox{0.8}{\,±\,5.4} & 69.4\scalebox{0.8}{\,±\,2.1} & 43.4\scalebox{0.8}{\,±\,6.4} & 60.8\scalebox{0.8}{\,±\,5.1} & 56.0\scalebox{0.8}{\,±\,5.1} & 67.1\scalebox{0.8}{\,±\,5.1} & 54.7\scalebox{0.8}{\,±\,2.1} & 59.8\scalebox{0.8}{\,±\,5.3} & 41.8\scalebox{0.8}{\,±\,6.7} & 46.4\scalebox{0.8}{\,±\,3.2} \\
    \cmidrule{1-1} \cmidrule(l{1.5mm}r{0mm}){2-7} \cmidrule(l){8-11}
    \texttt{OFA}~\scalebox{0.75}{~(\mycite{liu2023one}\textit{ICLR'24})}   & 58.0\scalebox{0.8}{\,±\,3.7} & 65.1\scalebox{0.8}{\,±\,3.9} & 45.4\scalebox{0.8}{\,±\,6.6} & 60.0\scalebox{0.8}{\,±\,6.6} & 59.7\scalebox{0.8}{\,±\,4.3} & 67.2\scalebox{0.8}{\,±\,3.4} & 56.0\scalebox{0.8}{\,±\,3.9} & 60.1\scalebox{0.8}{\,±\,3.5} & 42.4\scalebox{0.8}{\,±\,5.8} & 48.1\scalebox{0.8}{\,±\,2.2} \\
    \texttt{ZeroG}~\scalebox{0.75}{~(\mycite{li2024zerog}\textit{KDD'24})} & 65.1\scalebox{0.8}{\,±\,2.2} & 74.2\scalebox{0.8}{\,±\,1.6} & 50.3\scalebox{0.8}{\,±\,5.7} & \underline{64.0\scalebox{0.8}{\,±\,5.1}} & 61.4\scalebox{0.8}{\,±\,4.0} & 70.2\scalebox{0.8}{\,±\,1.3} & 58.5\scalebox{0.8}{\,±\,4.0} & \underline{66.2\scalebox{0.8}{\,±\,3.9}} & 46.1\scalebox{0.8}{\,±\,5.9} & 57.8\scalebox{0.8}{\,±\,3.6} \\
    \texttt{GraphCLIP}~\scalebox{0.75}{~(\mycite{zhu2024graphclip}\textit{WWW'25})} & 65.9\scalebox{0.8}{\,±\,3.7} & 75.1\scalebox{0.8}{\,±\,2.2} & 50.4\scalebox{0.8}{\,±\,4.0} & 63.0\scalebox{0.8}{\,±\,3.3} & 60.7\scalebox{0.8}{\,±\,3.8} & 71.3\scalebox{0.8}{\,±\,2.2} & \underline{58.6\scalebox{0.8}{\,±\,2.2}} & 65.7\scalebox{0.8}{\,±\,2.2} & 46.0\scalebox{0.8}{\,±\,3.3} & 58.8\scalebox{0.8}{\,±\,4.4} \\
    \texttt{UniGraph}~\scalebox{0.75}{~(\mycite{he2024unigraph}\textit{KDD'25})} & \underline{66.5\scalebox{0.8}{\,±\,2.5}} & \underline{76.5\scalebox{0.8}{\,±\,1.0}} & \underline{50.9\scalebox{0.8}{\,±\,4.4}} & 64.0\scalebox{0.8}{\,±\,2.4} & \underline{61.5\scalebox{0.8}{\,±\,2.6}} & \underline{71.4\scalebox{0.8}{\,±\,2.3}} & 58.1\scalebox{0.8}{\,±\,4.0} & 66.0\scalebox{0.8}{\,±\,3.8} & \underline{47.0\scalebox{0.8}{\,±\,2.5}} & \underline{58.9\scalebox{0.8}{\,±\,2.2}} \\
    \cmidrule{1-1} \cmidrule(l{1.5mm}r{0mm}){2-7} \cmidrule(l){8-11}
    \textbf{\modelname~\textit{(ours)}} & \textbf{68.7\scalebox{0.8}{\,±\,1.5}} & \textbf{78.4\scalebox{0.8}{\,±\,0.6}} & \textbf{52.2\scalebox{0.8}{\,±\,6.1}} & \textbf{65.5\scalebox{0.8}{\,±\,2.2}} & \textbf{62.4\scalebox{0.8}{\,±\,2.1}} & \textbf{71.5\scalebox{0.8}{\,±\,1.9}} & \textbf{60.2\scalebox{0.8}{\,±\,4.2}} & \textbf{68.0\scalebox{0.8}{\,±\,3.1}} & \textbf{48.1\scalebox{0.8}{\,±\,1.1}} & \textbf{62.0\scalebox{0.8}{\,±\,4.3}} \\
    \bottomrule
    \end{tabular}%
}
\label{tab:rq1}
\end{table*}

%% file: 6_conclusion.tex
\vspace{0.3cm}
\section{Conclusion}
\label{sec:conclusion}
In this work, we propose a Retrieval-Augmented Generation aided Graph Foundation Model named \modelname~that mitigates the in-memory bottleneck of existing GFMs by externalizing knowledge into a unified semantic-structural retrieval database. Instead of encoding priors into parameters, \modelname~decouples parameterized learning from retrievable knowledge, enabling interpretable and efficient adaptation. Through cross-view alignment and retrieval-augmented prompting, the framework achieves efficient generalization across domains and datasets. Extensive experiments demonstrate that \modelname~consistently surpasses state-of-the-art GFMs in effectiveness, efficiency, and robustness across diverse settings.

%% file: appendix/2_alg.tex
\section{Algorithms and Complexity Analysis}
\label{app:alg}
We illustrate the overall pre-training pipeline of \modelname~in Algorithm~\ref{alg:pretrain}, and the fine-tuning pipeline in Algorithm~\ref{alg:finetune}.

\subsection{Complexity Analsys of Algorithm~\ref{alg:pretrain}}
The pre-training pipeline mainly consists of three stages:

\textbf{Database Construction.}
For each source domain $D_i^{\mathcal S}$ with $N_i$ nodes, $E_i$ edges, and feature dimension $d_i$, we first project node features into a unified space of dimension $d_0$ via PCA, which costs $\mathcal{O}(N_i d_i d_0)$. BERT encodes each node’s text with the complexity of $\mathcal{O}(N_i C_{\text{BERT}})$, where $C_{\text{BERT}}$ denotes the per-sample encoding cost. 
For the structural store, computing $K$-order WSE requires $\mathcal{O}(KE_i)$ with sparse matrix multiplications, followed by top-$M_i$ anchor selection $\mathcal{O}(N_i \log N_i)$ and $h$-hop ego-subgraph extraction $\mathcal{O}(M_i \bar{d}_i^h)$ with $\bar{d}_i$ as average degree. 
Over $n$ domains, the complexity is:
\begin{equation}
    \mathcal{O}\Big(\sum_{i=1}^n \big[N_i d_i d_0 + N_i C_{\text{BERT}} + K E_i + N_i \log N_i + M_i \bar{d}_i^h \big]\Big).
\end{equation}

\textbf{Cross-view Encoding.}
At each iteration, a mixed-domain batch $\mathcal B$ of size $|\mathcal B|$ is sampled. Two $L$-layer GNN encoders with dimension $d$ are applied to semantic and structural views, giving complexity:
\begin{equation}
    \mathcal{O}\big(2 L (E_{\mathcal B}+|\mathcal B|) d \big),
\end{equation}
where $E_{\mathcal B}$ is the number of edges in the sampled batch $\mathcal B$.

\textbf{Self-supervised Pre-training.}
The cross-view InfoNCE computes pairwise similarities, with cost $\mathcal{O}(|\mathcal B|^2 d)$. The compression regularizers introduce $\mathcal{O}(|\mathcal B| d)$, negligible compared to the quadratic term. Token regularization across $n$ domains costs $\mathcal{O}(nd)$.

\textbf{Overall Complexity.}
The dominant cost per iteration is:
\begin{equation}
   \mathcal{O}\big(L (E_{\mathcal B}+|\mathcal B|) d + |\mathcal B|^2 d \big),
\end{equation}
while database construction is a one-time preprocessing overhead.

\textbf{Summary.}
The dominant cost comes from GNN propagation and quadratic contrastive alignment. Database construction is performed once and is negligible compared to iterative training.

\input{table/alg_pretrain}

\subsection{Complexity Analsys of Algorithm~\ref{alg:finetune}}
In the fine-tuning phase, the encoders $\boldsymbol{\Theta}_t^\star, \boldsymbol{\Theta}_s^\star$ and domain tokens $\{\boldsymbol{\tau}_{D}\}$ are frozen, and only prompt parameters $\boldsymbol{\Omega}$ are optimized.

\textbf{Preprocessing.}
For $m$-shot support instances with raw dimension $d^{\mathcal T}$, preprocessing requires $\mathcal{O}(m d^{\mathcal T} d_0 + m C_{\text{BERT}})$.

\textbf{Domain-gated Fusion.}
For each support instance, similarities with $n$ domain tokens are computed at cost $\mathcal{O}(mnd)$.

\textbf{Semantic Retrieval.}
Each query searches $\mathcal D_{\text{text}}$ of size $M_{\text{text}}$ using approximate nearest neighbor (ANN) with $\mathcal{O}(\log M_{\text{text}})$ per query. Aggregating top-$k$ answers incurs $\mathcal{O}(k d_0)$. The total cost is:
\begin{equation}
    \mathcal{O}\big(m (\log M_{\text{text}} + k d_0)\big).
\end{equation}

\textbf{Structural Retrieval.}
Each query searches the structural store of $n$ domains, each of size $M_{\text{struct}}$. ANN search costs $\mathcal{O}(n \cdot \log M_{\text{struct}})$, and fusing motif features requires $\mathcal{O}(n d)$. Thus:
\begin{equation}
    \mathcal{O}\big(m (n \cdot \log M_{\text{struct}} + n d)\big).
\end{equation}

\textbf{Prompted adaptation.}
Prompt construction and concatenation cost $\mathcal{O}(m d)$. The InfoNCE fine-tuning loss requires similarity with $C$ class prototypes, giving $\mathcal{O}(m C d)$.

\textbf{Overall complexity.}
The dominant cost per iteration is:
\begin{equation}
    \mathcal{O}\big(m [\log M_{\text{text}} + n \cdot \log M_{\text{struct}} + (k+n+C) d]\big).
\end{equation}

\textbf{Summary.}
The main bottleneck lies in retrieval (logarithmic in database size) and classification overhead. Since the backbone parameters are frozen and only lightweight prompts are updated, fine-tuning is substantially more efficient than full adaptation.

\input{table/alg_funetune}

%% file: table/alg_pretrain.tex
\begin{algorithm}[t]
\caption{Overall pre-training pipeline of \modelname.}
\label{alg:pretrain}
\KwIn{$n$ source graphs $\{G_i^{\mathcal{S}}\}_{i=1}^n$ from domain $\{D^{\mathcal{S}}\}$; Batch size $\mathcal{B}$; Learning rate $\eta_1$; Pre-training epochs $E_1$.}
\KwOut{Graph learner $h=g \circ f$ with parameters $\boldsymbol{\Theta}^\star_t$ and $\boldsymbol{\Theta}^\star_s$; Domain tokens $\{\boldsymbol{\tau}_{D_i}\}_{i=1}^n$.}
\BlankLine
Initialize all learnable parameters randomly\;
\textcolor{gray}{\tcp{Establish the Unified Retrieval Database}}
\For{each $G_i^\mathcal{S}$ in $\{G_i^{\mathcal{S}}\}_{i=1}^n$}{
Representation track: $\mathbf{Z}_i^\mathcal S \gets$~Eq.~\eqref{eq:align}, Eq.~\eqref{eq:semantic_view}\;
Retrieval track: $\widetilde{\mathbf{z}}_{v}^\mathcal S\gets$~Eq.~\eqref{eq:retrieval_track} for each node $v$\;
Establish the semantic store: $\mathcal{D}_{\text{text}}\gets$~Eq.~\eqref{eq:retrieval_track}\;
Establish the structural store: $\mathcal{D}_{\text{struct}}\gets$~Eq.~\eqref{eq:structual_store}\;
}
\textcolor{gray}{\tcp{Compose Node Views}}
\For{$e_1=1,2,\cdots, E_1$}{
\For{each $G_i^\mathcal{S}$ in $\{G_i^{\mathcal{S}}\}_{i=1}^n$}{
Learn node semantic view: $\mathbf{Z}_i^\mathcal S \gets$~Eq.~\eqref{eq:align}, Eq.~\eqref{eq:semantic_view}\;
Learn node strcutural view: $\mathbf{W}_i^\mathcal S \gets$~Eq.~\eqref{eq:strcutural_view}\;
\textcolor{gray}{\tcp{Self-supervised Information Bottleneck}}
Encode dual-embeddings: $\mathbf{H}_i^\text{text}, \mathbf{H}_i^\text{struct}\gets$~Eq.~\eqref{eq:pretrain_encode}\;
\For{each node $v$ in the sampled batch $\mathcal{B}$}{
Calculate the alignment loss: $\mathcal{L}_{\text{align}}^{(i,v)}\gets$~Eq.~\eqref{eq:intractbale}\;
}
}
\textcolor{gray}{\tcp{Token Regularization and Parameter Update}}
Calculate overall pre-training loss: $\mathcal{L}_{\text{pretrain}}\gets$~Eq.~\eqref{eq:pretrain_loss}\;
Update the model parameters $\boldsymbol{\Theta}_t,\boldsymbol{\Theta}_s$ by minimizing $\mathcal L_{\text{pretrain}}$ and back-propagation with learning rate $\eta_1$\;
}
\end{algorithm}

%% file: table/alg_funetune.tex
\begin{algorithm}[t]
\caption{Overall fine-tuning pipeline of \modelname.}
\label{alg:finetune}
\KwIn{Unified database $\mathcal{D}$; Target domain $D^\mathcal{T}$; Target graph(s) and $m$-shot support set $\mathcal{S}^\mathcal{T}$; Frozen parameters $\boldsymbol{\Theta}^\star_t$ and $\boldsymbol{\Theta}^\star_s$; Frozen domain tokens $\{\boldsymbol{\tau}_{D_i}\}_{i=1}^n$; Learning rate $\eta_2$; Fine-tuning epochs $E_2$.}
\KwOut{Fine-tuned graph learner $h^\star=g^\star(f^\star)$ with parameters $\{\boldsymbol{\Theta}^\star_t, \boldsymbol{\Theta}^\star_s, \boldsymbol{\Omega}^\star\}$.}
\BlankLine
Initialize all learnable parameters randomly\;
\textcolor{gray}{\tcp{Preprocess Support Set}}
\For{each support node (or graph) in $\mathcal{S}^\mathcal{T}$}{
Learn dimension-aligned feature $\widehat{\mathbf{X}}^\mathcal{T}\gets$~Eq.~\eqref{eq:align}, Eq.~\eqref{eq:semantic_view}\;
}
\For{$e_1=1,2,\cdots, E_2$}{
\For{each support node (or graph) in $\mathcal{S}^\mathcal{T}$}{
Encode via pre-trained learner: $\mathbf{Z}_i^\mathcal T\gets$~Eq.~\eqref{eq:gate}\;
\textcolor{gray}{\tcp{Domain-gated Fusion}}
Calculate gating weights: $\{\pi_{i,k}\}_{k=1}^n\gets$~Eq.~\eqref{eq:gate}\;
\textcolor{gray}{\tcp{Semantic Query and Retrieval}}
Query $\mathbf{D}_\text{text}$ and get answers: $\big(\Delta\mathbf{z}^{\mathcal{T}}_v\big)^\text{text} \gets$~Eq.~\eqref{eq:text_query}\;
Query $\mathbf{D}_\text{struct}$ and get answers: $\big(\Delta\mathbf{z}^{\mathcal{T}}_v\big)^\text{struct} \!\!\gets$~Eq.~\eqref{eq:struct_query}\;
\textcolor{gray}{\tcp{In-context Augmentation and Prompt}}
Update instance embedding: $\mathbf{z}_v^\mathcal{T\prime\prime}\!\!\gets$~Eq.~\eqref{eq:augment_1}, Eq.~\eqref{eq:augment_2}\;
Inialize $\boldsymbol{P_\Omega}\gets$~Eq.~\eqref{eq:prompt} and prompt: $\mathbf{h}_i^\mathcal{T}\gets$~Eq.~\eqref{eq:prompt}\;
}
\textcolor{gray}{\tcp{Few-shot Adaptation and Parameter Update}}
Calculate the fine-tuning loss: $\mathcal{L}_{\text{fine-tne}}\gets$~Eq.~\eqref{eq:finetune_loss}\;
Update the prompt parameters $\boldsymbol{\Omega}$ by minimizing $\mathcal L_{\text{fine-tne}}$ and back-propagation with learning rate $\eta_2$\;
}
\end{algorithm}

%% file: appendix/3_proof.tex
\section{Proofs}
\label{app:proof}
\subsection{Proof of Proposition~\ref{prop:wse}}
\label{app:proof1}
We first restate Proposition~\ref{prop:wse} for reference.
\begin{mathbox}
    \textbf{Proposition 1} (Structural Separability of WSE)\textbf{.}
        There exist pairs of non-isomorphic graphs $G_1$, $G_2$ and nodes $v\in G_1$, $u\in G_2$ such that for any fixed radius $r$, the $r$-hop neighbors $\mathcal{N}_r(v)$ and $\mathcal{N}_r(u)$ are isomorphic, yet the Walk-Spectrum Encodings satisfy:
        \begin{equation}
            \mathbf{C}^{\text{WSE}}_\alpha(v) \ne \mathbf{C}^{\text{WSE}}_\alpha(u).
        \notag
        \end{equation}
\end{mathbox}
\begin{proof}
    The sketch is to construct a pair of graphs that are cospectral locally (same $r$-neighbor) but differ in global cycle structure (\eg, attaching different length cycles far from the root while preserving the first $r$ shells). Closed-walk counts at the root incorporate returns that traverse those distant cycles, which appear only at higher orders. Hence, a finite $K$ separates $v$ and $u$. 
    This ensures that traditional $r$-hop methods~\cite{southern2025balancing} cannot distinguish them, while WSE produces different signatures.

    Formally, fix any radius $r$. Let $P_{r+1} = (x_0, x_1, \ldots, x_{r+1})$ be a path of length $r+1$, with root $x_0$. At the endpoint $x_{r+1}$, we attach a cycle.
    
    In graph $G_1$, we attach an odd cycle $C_p$ of length $p \geqslant 3$, while in graph $G_2$, we attach an even cycle $C_q$ of length $q \geqslant 4$. Denote the roots by $v=x_0 \in G_1$ and $u=x_0 \in G_2$.
    
    As the cycle appears only beyond radius $r$, the neighbors $\mathcal{N}_r(v)$ and $\mathcal{N}_r(u)$ both reduce to path $(x_0,\cdots,x_r)$, and are isomorphic:
    \begin{equation}
        \mathcal{N}_r(v) \cong \mathcal{N}_r(u).
    \end{equation}
    Consider closed walks that go from the root to the cycle, traverse it, and return. In $G_1$, the shortest such closed walk has length $K_1 = 2(r+1) + p$, while in $G_2$ the shortest length is $K_2 = 2(r+1) + q$, and in fact any closed walk using the cycle in $G_2$ has length:
    \begin{equation}
        K = 2(r+1) + q + 2\ell, \quad \ell \geqslant 0,
    \end{equation}
    which is always even. Since $p$ is odd,$ K_1$ is odd, and therefore:
    \begin{equation}
        \big(\mathbf{A}_{G_1}^{K_1}\big)_{vv} > 0, \quad \big(\mathbf{A}_{G_2}^{K_1}\big)_{uu} = 0.
    \end{equation}
    By the definition of WSE that $C^{\mathrm{WSE}}_{\alpha}(z)[k] = \alpha^k \mathbf{A}^k_{zz}$, so the two encodings must differ at coordinate $K_1$. Hence, we conclude that:
    \begin{equation}
        \mathbf{C}^{\text{WSE}}_\alpha(v) \ne \mathbf{C}^{\text{WSE}}_\alpha(u),
    \end{equation}
    which establishes that WSE separates nodes indistinguishable by local neighborhoods. We conclude the proof.
\end{proof}

\subsection{Proof of Proposition~\ref{prop:ib}}
\label{app:proof2}
We first restate Proposition~\ref{prop:ib} for reference.
\begin{mathbox}
    \textbf{Proposition 2} (Cross-View Mutual Information Bounds)\textbf{.}
        The relevance term admits the InfoNCE lower-bound estimator:
        \begin{equation}
            \!\!\!\!\!\!I\big(\mathbf{h}^{\text{text}}_{i,v};\mathbf{h}^{\text{struct}}_{i,v}\big) \!\leqslant\! \frac{1}{|\mathcal{B}|}\!\sum_{v\in\mathcal{B}}\!\log\!\frac{\exp\!\big(\sigma\big(g_t(\mathbf{h}^{\text{text}}_{i,v}),g_s(\mathbf{h}^{\text{struct}}_{i,v})\big)\!/\!\tau\big)}{\sum\limits_{u\in\mathcal{B}}\!\!\exp\!\big(\sigma\big(g_t(\mathbf{h}^{\text{text}}_{i,v}),g_s(\mathbf{h}^{\text{struct}}_{i,u})\big)\!/\!\tau\big)},\!\!\!\!
        \notag
        \end{equation}
        where $g_t$, $ g_s$ are projections, $\sigma(\cdot)$ is similarity, $\tau$ is a temperature, positives are formed by the same node across the views $(v,v)$ in a batch $\mathcal{B}$, and negatives by mismatched nodes $(v,u), u \!\neq\! v$.
        \vspace{0.1cm}
        \noindent\makebox[\linewidth]{\dotfill}
        The compression term can be upper-bounded via KL-divergence:
        \begin{equation}
            \!\!\!\!I\big(\mathbf{h}_{i,v}^\cdot;\overline{\mathbf{x}}~\!_{i,v}^{\mathcal S}\big) \!\!\geqslant\! \!\mathbb{E}_{p(\mathbf{h}_{i,v}^\cdot,\overline{\mathbf{x}}~\!_{i,v}^{\mathcal S})} \!\big[\!\log q_\phi\big(\mathbf{h}_{i,v}^\cdot|\overline{\mathbf{x}}~\!_{i,v}^{\mathcal S}\big)\!\big] \!-\! \mathbb{E}_{p(\mathbf{h}_{i,v}^\cdot)}\! \big[\!\log p\big(\mathbf{h}_{i,v}^\cdot\big)\!\big],\!\!
        \notag
        \end{equation}
        where $v$ is sampled from $\mathcal{B}$, $\mathbf{x}$ denotes $\mathbf{z}$ or $\mathbf{w}$, and $q_\phi(\cdot|\cdot)$ is a variational approximation of the true conditional distribution.
\end{mathbox}
\begin{proof}
    
    \textbf{(1) Relevance Term} (InfoNCE Lower Bound)\textbf{.}
    The mutual information between semantic and structural embeddings is:
    \begin{equation}
        I(\mathbf{h}^{\text{text}}_{i,v}; \mathbf{h}^{\text{struct}}_{i,v}) = \mathbb{E}_{p(\mathbf{h}^{\text{text}}_{i,v}, \mathbf{h}^{\text{struct}}_{i,v})}\left[\log\frac{p(\mathbf{h}^{\text{text}}_{i,v}| \mathbf{h}^{\text{struct}}_{i,v})}{p(\mathbf{h}^{\text{text}}_{i,v})}\right].
    \end{equation}
    
    Directly computing is intractable. Following the contrastive estimation framework of InfoNCE~\cite{oord2018representation}, we approximate it with a similarity-based classification task that distinguishes positive pairs (same node across views) from negative pairs (different nodes).
    
    Given a batch $\mathcal{B}$ of nodes, we define the similarity score $s_{vu}=\sigma(g_t(\mathbf{h}^{\text{text}}_{i,v}), g_s(\mathbf{h}^{\text{struct}}_{i,u})),$ where $g_t$, $g_s$ are projection heads and $\sigma(\cdot)$ is similarity. Then the InfoNCE lower bound becomes:
    \begin{equation}
        I(\mathbf{h}^{\text{text}}_{i,v}; \mathbf{h}^{\text{struct}}_{i,v}) \geqslant \frac{1}{|\mathcal{B}|} \sum_{v\in\mathcal{B}} \log\frac{\exp(s_{vv}/\tau)}{\sum_{u\in\mathcal{B}} \exp(s_{vu}/\tau)},
    \end{equation}
    where $\tau$ is a temperature. This corresponds to Eq.~\eqref{eq:lb} and provides a variational lower bound of the cross-view relevance, encouraging semantic and structural embeddings of the same node to align while contrasting mismatched pairs.

    \textbf{(2) Compression Term} (KL Upper Bound)\textbf{.}
    We consider the mutual information between an embedding $\mathbf{h}_{i,v}^\cdot$ (either semantic or structural) and its input feature $\mathbf{X}^{S}_{i}$. By definition:
    \begin{align}
        I(\mathbf{h}_{i,v}^\cdot; \mathbf{X}^{S}_{i})&= \mathbb{E}_{p(\mathbf{h}, \mathbf{X})}\left[\log \frac{p(\mathbf{h}|\mathbf{X})}{p(\mathbf{h})}\right]=\mathbb{E}_{p(\mathbf{h}, \mathbf{X})}[\log p(\mathbf{h}|\mathbf{X})]\notag\\
        &=\mathbb{E}_{p(\mathbf{h})}[\log p(\mathbf{h})].
    \label{eq:temp9}
    \end{align}
    Since the conditional posterior $p(\mathbf{h}|\mathbf{X})$ is unknown, we introduce a variational approximation $q_{\phi}(\mathbf{h}|\mathbf{X})$. Using the non-negativity of the KL divergence, we obtain an upper bound:
    \begin{equation}
        \mathbb{E}_{p(\mathbf{h},\mathbf{X})}[\log p(\mathbf{h}|\mathbf{X})] \leqslant \mathbb{E}_{p(\mathbf{h},\mathbf{X})}[\log q_{\phi}(\mathbf{h}|\mathbf{X})].
    \label{eq:temp10}
    \end{equation}
    Substituting Eq.~\eqref{eq:temp10} into Eq.~\eqref{eq:temp9} gives:
    \begin{equation}
        I(\mathbf{h}_{i,v}^\cdot; \mathbf{X}^{S}_{i}) \leqslant \mathbb{E}_{p(\mathbf{h},\mathbf{X})}[\log q_{\phi}(\mathbf{h}|\mathbf{X})]=\mathbb{E}_{p(\mathbf{h})}[\log p(\mathbf{h})],
    \end{equation}
    which corresponds to Eq.~\eqref{eq:ub}. This upper bound serves as a variational surrogate that penalizes redundant signals while maintaining tractability. The first term encourages compression through reconstruction under $q_{\phi}$, and the second term regularizes the marginal entropy of the latent representation.
\end{proof}

%% file: appendix/4_exp_settings.tex
\section{Experiment Details}
\label{app:exp_detail}
\input{table/dataset}

\subsection{Dataset Details}
\label{app:exp_detail_data}
\begin{itemize}[leftmargin=*]
    \item \textbf{Citation Domain:} \texttt{Cora}~\cite{mccallum2000automating} , \texttt{CiteSeer}~\cite{giles1998citeseer}, and \texttt{PubMed}~\cite{sen2008collective}, where nodes represent papers and edges denote citation links. Each node is equipped with text-based features derived from titles or abstracts.
    \item \textbf{E-Commerce Domain:} contains two subgraphs from the large-scale \texttt{Ogbn-Products}~\cite{hu2020open}, including \texttt{Ogbn-Tech} and \texttt{Ogbn-Home}. Nodes represent Amazon products, edges indicate co-purchase relationships, and node labels correspond to product categories, capturing consumer behavior.
    \item \textbf{Web Link Domain:} consists of the \texttt{Wiki-CS}~\cite{mernyei2020wiki} dataset, where nodes correspond to Wikipedia articles and edges represent hyperlinks. Textual embeddings extracted from article content provide rich semantic information for web-scale graph learning.
\end{itemize}

\subsection{Implementation Details}
We introduce the general implementation details below. 

\textbf{Pre-training.}
We pre-train \modelname~for up to 10,000 epochs with early stopping for 50 consecutive epochs. Both semantic and structural encoders are 2-layer GNNs. The overall pre-training objective combines the cross-view information bottleneck loss and the domain-token regularizer weighted by $\gamma$ tuned within the range of $[0,1]$. The Adam optimizer is adopted, with the learning rate and weight decay selected from [10$^{-5}$, 10$^{-1}$] via grid search on the validation set. All parameters are initialized from scratch.

\textbf{Fine-tuning.}
We fine-tune the \modelname~for up to 100 episodes with an early stopping strategy. The pretrained encoder parameters are frozen, and only the prompt parameters are updated. For each query node, we retrieve $k$ ($[1,10]$) query-answer pairs from both the semantic and structural databases, which are fused with weights $\lambda_{\text{text}}$ ($[0,1]$) and $\lambda_{\text{struct}}$ ($[0,1]$) to form retrieval-augmented prompts. The final fine-tuning objective is optimized by Adam with the same learning rate and weight decay settings as in pre-training.

\textbf{Environment.}
Experiments are conducted with:
\begin{itemize}[leftmargin=*]
    \item Operating System: Ubuntu 20.04 LTS.
    \item CPU: Intel(R) Xeon(R) Platinum 8358 CPU@2.60GHz with 1TB DDR4 of Memory.
    \item GPU: NVIDIA Tesla A100 SMX4 with 80GB of Memory.
    \item Software: CUDA 10.1, Python 3.8.12, PyTorch  1.9.1, PyTorch Geometric 2.0.1, NanoVectorDB 0.0.4.3.
\end{itemize}

%% file: table/dataset.tex
\begin{table}[!t]
  \centering
  \setlength{\tabcolsep}{2pt}
  \caption{Statistics of the multi-domain graph dataset.}
  \vspace{-0.1cm}
  \resizebox{0.48\textwidth}{!}{
    \begin{tabular}{llrrrrr}
    \toprule
    \textbf{Dataset} & \textbf{Domain} & $\#$\textbf{Node} & $\#$\textbf{Edge} & \makecell[r]{$\#$\textbf{Feat.}\\\textbf{Dim.}} & $\#$\textbf{Class} & \makecell[r]{\textbf{Avg.}\\$\#$\textbf{Deg.}} \\
    \midrule
    \texttt{Cora}~\cite{mccallum2000automating}  & Citation & 2,708 & 5,429 & 1,433 & 7     & 4.00\\
    \texttt{CiteSeer}~\cite{giles1998citeseer} & Citation & 3,186 & 4,277 & 3,703 & 6     & 2.57 \\
    \texttt{PubMed}~\cite{sen2008collective} & Citation & 19,717 & 44,338 & 500   & 3     & 4.50 \\
    \midrule
    \makecell[l]{\texttt{Ogbn-Products}\\{\small(Tech.)}~\cite{hu2020open}} & E-Commerce & 47,428 & 2,077,241 & 100   & 3    & 87.60 \\
    \makecell[l]{\texttt{Ogbn-Products}\\{\small(Home)}~\cite{hu2020open}} & E-Commerce & 9,790 & 131,841 & 100   & 5     & 26.93 \\
    \midrule
    \texttt{Wiki-CS}~\cite{mernyei2020wiki} & Web Link & 11,701 & 216,123 & 300   & 10    & 36.94 \\
    \bottomrule
    \end{tabular}%
    }
  \label{tab:dataset}%
\end{table}%